\documentclass[sigconf, balance=false]{acmart}

\usepackage{hyperref}       
\usepackage{url}            
\usepackage{booktabs}       
\usepackage{amsfonts}       
\usepackage{microtype}      

\usepackage{amsmath, amsthm, bbm, mathtools, commath}
\usepackage{datetime}
\usepackage{hyperref}
\usepackage{verbatim}
\usepackage[font=small]{subfig}
\usepackage{colortbl}
\usepackage{arydshln} 
\graphicspath{ {./visualizations/} }

\newcommand{\mbf}{\mathbf}

\newcommand{\bestcell}{\cellcolor{blue!25}}
\newcommand{\myparagraph}[1]{\smallskip \noindent \textbf{#1}}

\newcommand{\std}[1]{\pm #1}

\AtBeginDocument{%
  \providecommand\BibTeX{{%
    \normalfont B\kern-0.5em{\scshape i\kern-0.25em b}\kern-0.8em\TeX}}}

\setcopyright{rightsretained}
\copyrightyear{2021}
\acmYear{2021}
\acmDOI{}

\acmConference[WWW '21]{WWW '21: The Web Conference}{April 19--23, 2021}{Ljubljana, Slovenia.}
\acmBooktitle{Workshop on Graph Learning Benchmarks, WWW '21,
  April 19--23, 2021, Ljubljana, Slovenia.}
\acmPrice{}
\acmISBN{}

\acmSubmissionID{}

\begin{document}

\title{New Benchmarks for Learning on Non-Homophilous Graphs}

\author{Derek Lim}
\authornote{Authors contributed equally to this research.}
\email{dl772@cornell.edu}
\orcid{0000-0001-8408-9484} 
\affiliation{%
  \institution{Cornell University}
  \country{USA}
}

\author{Xiuyu Li}
\authornotemark[1]
\email{xl289@cornell.edu}
\affiliation{%
  \institution{Cornell University}
  \country{USA}
}

\author{Felix Hohne}
\authornotemark[1]
\email{fmh42@cornell.edu}
\affiliation{%
  \institution{Cornell University}
  \country{USA}
  }
  
\author{Ser-Nam Lim}
\email{sernam@gmail.com}
\affiliation{%
\institution{Facebook AI}
\country{USA}
}


\begin{abstract}
Much data with graph structures satisfy the principle of homophily, meaning that connected nodes tend to be similar with respect to a specific attribute. As such, ubiquitous datasets for graph machine learning tasks have generally been highly homophilous, rewarding methods that leverage homophily as an inductive bias. Recent work has pointed out this particular focus, as new non-homophilous datasets have been introduced and graph representation learning models better suited for low-homophily settings have been developed. However, these datasets are small and poorly suited to truly testing the effectiveness of new methods in non-homophilous settings. We present a series of improved graph datasets with node label relationships that do not satisfy the homophily principle. Along with this, we introduce a new measure of the presence or absence of homophily that is better suited than existing measures in different regimes. We benchmark a range of simple methods and graph neural networks across our proposed datasets, drawing new insights for further research. Data and codes can be found at \href{https://github.com/CUAI/Non-Homophily-Benchmarks}{\color[HTML]{395dc0}https://github.com/CUAI/Non-Homophily-Benchmarks}.
\end{abstract}




\maketitle

\section{Introduction}

Various types of real-world data have natural graph structures, in which objects (nodes) and relationships (edges) are encoded as a graph $G$. As a result, numerous types of models have been developed for machine learning tasks on graph data. Much recent work in graph representation learning \cite{hamiltonbook} has focused on using both node features and graph structure, especially in the general family of graph neural network (GNN) models.

To generate predictions, graph learning methods can leverage complex inductive biases relating to the topology of the graph \cite{battaglia2018relational}. One common property of real-world graphs that is often leveraged as a strong inductive bias is \textit{homophily}, which means that connected nodes tend to be similar in certain attributes \cite{mcpherson2001birds, altenburger2018monophily}.
Here, we refer to a graph as \textit{homophilous} if connected nodes are much more likely to have the same class label than if edges were independent of labels.
In a \textit{non-homophilous graph}, connected nodes are not significantly more likely to have the same class label, and may even be less likely to have the same class label than if edges were independent of labels.
For instance, nodes may have little to no preference for connecting to any particular class in aggregate (e.g. in friendship networks where the classes represent gender \cite{laniado2016gender, altenburger2018monophily}); or, nodes of a certain class may only be connected to a subset of the other classes (e.g. in citation networks where class labels are the year of publication of a paper).

Many GNNs explicitly assume homophily in their construction, and even those models that do not may perform poorly in non-homophilous graph settings \cite{zhu2020beyond, jia2020residual}. While some GNNs have been developed that work better in non-homophilous settings, their evaluation is often limited to a few graph datasets introduced by \citet{pei2019geom} that have certain undesirable properties such as small size, narrow range of application areas, and synthetic classes \cite{zhu2020beyond, nonlocal, zhu2020graph, chien2021adaptive, chen2020simple, yan2021two}.

In this work, we summarize issues with the non-homophilous graph datasets that are used in prior work and propose improved non-homophilous graph datasets that are substantially larger, span wider application areas, and capture different types of complex label-topology relationships.
As previously-used homophily metrics have various shortcomings, we introduce a new metric that detects the presence or absence of homophily in graphs, and use it to analyze these graph datasets.
We reintroduce strong simple methods for learning on graphs that have been overlooked in recent non-homophilous GNN work. Finally, we benchmark these simple methods along with general GNNs and non-homophilous GNNs on our proposed datasets. Our stronger empirical evaluation setup allows for a better understanding of graph learning methods in diverse settings.

\section{Prior Work}

\subsection{Graph Representation Learning}

Graph neural networks \cite{hamilton2017inductive, kipf2017semi, velivckovic2018graph} have demonstrated their utility on a variety of graph machine learning tasks. Most GNNs are constructed by stacking graph neural network layers that propagate transformed node features and then aggregate them via different mechanisms. The neighborhood aggregation used in many existing GNNs implicitly leverage homophily, so they often fail to generalize on non-homophilous graphs \cite{zhu2020beyond, balcilar2021analyzing}. Indeed, a wide range of GNNs operate as low-pass graph filters \cite{nt2019revisiting, wu2019simplifying, balcilar2021analyzing} that smooth features over the graph topology, which produces similar representations and thus similar predictions for neighboring nodes. 

Due in part to the most common graph learning benchmarks exhibiting strong homophily, various graph representation learning methods have been developed that explicitly make use of an assumption of homophily in the data \cite{wu2019simplifying, huang2021combining, deng2020graphzoom, klicpera2019diffusion, bojchevski2020scaling}. By leveraging this assumption, several simple, inexpensive models are able to achieve state-of-the-art performance on homophilic datasets \cite{wu2019simplifying, huang2021combining}.

\subsection{Non-Homophilous Methods}

Various GNNs have been proposed to achieve higher performance in low-homophily settings. For instance, Geom-GCN \cite{pei2019geom} introduces a geometric aggregation scheme, MixHop \cite{abu2019mixhop} uses a graph convolutional layer that mixes powers of the adjacency matrix, GPR-GNN \cite{chien2021adaptive} uses learnable weights that can be positive and negative in feature propagation, and H\textsubscript{2}GCN \cite{zhu2020beyond} shows that separation of ego and neighbor embeddings, aggregation in higher-order neighborhoods, and the combination of intermediate representations improves GNN performance in low-homophily. Also, various methods that only depend on graph topology have been proposed for non-homophilous settings, which are based on label propagation or supervised learning models \cite{peel2017graph, altenburger2018monophily, chin2019decoupled, zheleva2009to}. There are several recurring design decisions across these methods that appear to strengthen performance in non-homophilous settings: using higher-order neighborhoods, decoupling neighbor information from ego information, and combining graph information at different scales.

\subsection{Real-world Datasets}

Recently, the Open Graph Benchmark \cite{hu2020open} has provided a series of datasets and leaderboards that improve the quality of evaluation in graph representation learning; however, most of the node classification datasets tend to be homophilous, as noted in past work \cite{zhu2020beyond} and expanded upon in Appendix~\ref{sec:homophilous_stats}. A comparable set of high-quality benchmarks to evaluate non-homophilous methods does not currently exist. 

The most widely used datasets to evaluate non-homophilous graph representation learning methods were presented by \cite{pei2019geom} (see our Appendix Table \ref{tab:geom_gcn}); however, these datasets have fundamental issues. First, they are very small --- the Cornell, Texas, and Wisconsin datasets have between 180-250 nodes, and the largest of these datasets has 7,600 nodes. In analogy to certain pitfalls of graph neural network evaluation on small (homophilic) datasets discussed in \cite{shchur2018pitfalls}, evaluation on the datasets in \cite{pei2019geom} is plagued by high variance across different train/test splits (see results in \cite{zhu2020beyond}). Also, the small size of these datasets may make models more prone to overfitting \cite{dwivedi2020benchmarking}, and does not allow for experiments in scalability of GNNs  designed for performance in non-homophilous settings. As a result, they are not satisfactory for evaluating the performance of node classification models for non-homophilous graphs, and larger, real-world datasets are necessary. 

\citet{peel2017graph} also studies node classification on network datasets with various types of relationships between edges and labels. However, they only study methods that act on node labels, and thus their datasets do not necessarily have node features. We take inspiration from their work, by testing on Pokec and Facebook networks with node features that we define, and by introducing other year-prediction tasks on citation networks that have node features.

\subsection{Synthetic Data}

Synthetically generated or synthetically modified non-homophilous graph data may also be used for evaluation of graph machine learning methods. Past works have taken various approaches to this \cite{karimi2018homophily, abu2019mixhop, zhu2020beyond, kim2021how, chien2021adaptive}.  When generating the graph, they control homophily through modifications of some generative models over graphs \cite{barabasi1999emergence, abbe2017community, deshpande2018contextual}. They tend to either sample node feature vectors from a real graph or draw them from multivariate Gaussian distributions. While synthetic data are useful for investigating properties of models in controlled settings, we focus on benchmarks from real-world data that exhibit diverse types of complex structure.

\section{Datasets and Metrics}

\subsection{Measuring Homophily}\label{sec:measure}

Various metrics have been proposed to measure the homophily of a graph in a single scalar quantity. However, these metrics are sensitive to the number of classes and the number of nodes in each class.
Thus, we introduce a metric that better captures the presence or absence of homophily. Our metric does not distinguish between different non-homophilous settings (such as heterophily or independent edges); we argue that there are too many degrees of freedom in non-homophilous settings for a single scalar quantity to be able to distinguish them all.

Let $G = (V, E)$ be a graph with $n$ nodes, none of which are isolated. Further let each node $x \in V$ have a class label $k_x \in \{0, 1, \ldots, C-1\}$ for some number of classes $C$, and denote by $C_k$ the set of nodes in class $k$. In recent non-homophilous graph representation learning work, the edge homophily \cite{zhu2020beyond} has been defined as the proportion of edges that connect two nodes of the same class:
\begin{equation}\label{eq:edge_hom}
    h = \frac{| \{(x,v) \in E : k_x = k_v \} |}{|E|}.
\end{equation}
Another related measure is what we call the node homophily \cite{pei2019geom}, defined as
 $ \frac{1}{|V|} \sum_{x \in V} \frac{d_x^{(k_x)}}{d_x}$, in which $d_x$ is the number of neighbors of node $x$, and $d_x^{(k_x)}$ is the number of neighbors of $x$ that have the same class label. We focus on the edge homophily \eqref{eq:edge_hom} in this work, but find that node homophily tends to have similar qualitative behavior in experiments. 
 
The sensitivity of edge homophily to the number of classes and size of each class limits its utility. Note that if edges were rewired randomly independently of node labels, a node $x \in V$ would be expected to have $d_x^{(k_x)}/d_x \approx |C_k|/n$ as the proportion of nodes of the same class that they connect to \cite{altenburger2018monophily}. For a dataset with $C$ balanced classes, we would thus expect the edge homophily to be around $\frac{1}{C}$, so the interpretation of the measure depends on the number of classes. Also, if classes are imbalanced, then the edge homophily may be misleadingly large. For instance, if 99\% of nodes were of one class, then most edges would likely be within that same class, so the edge homophily would be high. 

We define a homophily measure that alleviates these shortcomings. Our measure is given as:
\begin{equation}\label{eq:our_measure}
\hat h = \frac{1}{C-1} \sum_{k=0}^{C-1} \left[h_k - \frac{|C_k|}{n} \right]_+,
\end{equation}
where $[a]_+ = \max(a, 0)$, and $h_k$ is the class-wise homophily metric
\begin{equation}
h_k = \frac{\sum_{x \in C_k} d_x^{(k_x)}}{\sum_{x \in C_k} d_x}.
\end{equation}

Note that $\hat h \in [0,1]$, with a fully homophilous graph (in which every node is only connected to nodes of the same class) having $\hat h = 1$. Since each class-wise homophily metric $h_k$ only contributes positive deviations from the null expected proportion $|C_k|/n$, the class-imbalance problem is substantially mitigated. Also, graphs in which edges are independent of node labels are expected to have $\hat h \approx 0$, for any number of classes. Our measure $\hat h$ measures presence of homophily, but does not distinguish between the many types of possibly non-homophilous relationships. This is reasonable given the diversity of non-homophilous relationships. For example, non-homophily can imply independence of edges and classes, extreme heterophily, connections only among subsets of classes, or certain chemically / biologically determined relationships. Indeed, these relationships are very different, and are better captured by more than one scalar quantity, such as the compatibility matrices that are discussed in Appendix~\ref{sec:appendix_measures}. Figure~\ref{fig:example_graphs} compares our measure $\hat h$ with the edge homophily ratio $h$.

On certain datasets where previous measures are misleading, our measure shows its advantages. For example, some of our proposed datasets are class-imbalanced (e.g. YelpChi and ogbn-proteins), so they have high edge homophily, but our measure $\hat h$ captures that they are indeed non-homophilous (Table \ref{tab:data_stats}). Moreover, as seen in Appendix Table \ref{tab:geom_gcn}, the edge homophily of Chameleon, Actor, and Squirrel are approximately the same, but the graph structures (Appendix Figure \ref{fig:compat_geom_gcn}) and performance of different methods on these datasets vary significantly \cite{zhu2020beyond}. According to our measure, Chameleon is more homophilous than Squirrel, which is in turn more homophilous than Actor, and past work has shown that models tend to perform better on Chameleon than Squirrel and better on Squirrel than Actor \cite{zhu2020beyond, chien2021adaptive, pei2019geom}. Our measure suggests one possible axis of variation that may help explain this divergence, but we emphasize that there are many possible confounders.
Further discussion is given in Appendix~\ref{sec:appendix_measures}.

\begin{table*}[ht]
    \centering
    \caption{Statistics of our proposed non-homophilous graph datasets. \# C is the number of distinct node classes. \# G/T is the number of different graphs or tasks.  When there are multiple graphs or tasks, the minimum and maximum statistics are listed, with a hyphen ``-'' in between.}
    \label{tab:data_stats}
    {\small
    \begin{tabular}{crrrrrrrr}
    \toprule
    Dataset & \# G/T & \# Nodes & \# Edges &  \# Node Feat. & \# C & Class types & Edge hom. & $\hat h$ (ours) \\
    \midrule
         Twitch-explicit & 7  & 1,912 - 9,498 & 31,299 - 153,138 & 2,545 & 2 & explicit content  & .556 - .632 & .049-.146 \\
         YelpChi & 1 & 45,954 & 3,846,979 & 32 & 2 & fake reviews & .773 & .052 \\
         deezer-europe & 1 & 28,281 & 92,752 & 31,241 & 2 & gender & .525 & .030 \\
         FB100 & 100 & 769-41,536 & 16,656-1,590,655 & 5 & 2 & gender &  .434 - .917 & .000 - .246  \\
         Pokec & 1 & 1,632,803 & 30,622,564 & 65  &  2 & gender & .445 & .000 \\
         ogbn-proteins & 112 & 132,534 & 39,561,252 & 8 & 2 & function & .623 - .940 & .090 - .260  \\
         arXiv-year & 1 & 169,343 & 1,166,243 & 128 & 5 & pub year & .222 & .272  \\
         snap-patents & 1 & 2,923,922 & 13,975,788 & 269 & 5 & time granted & .073 & .100  \\
    \bottomrule
    \end{tabular}
    }
\end{table*}

\begin{figure}
    \centering
    \begin{tabular}{cc}
    \subfloat[\normalsize $h=1$, \  $\hat h = 1$]{\includegraphics[width=.20\textwidth]{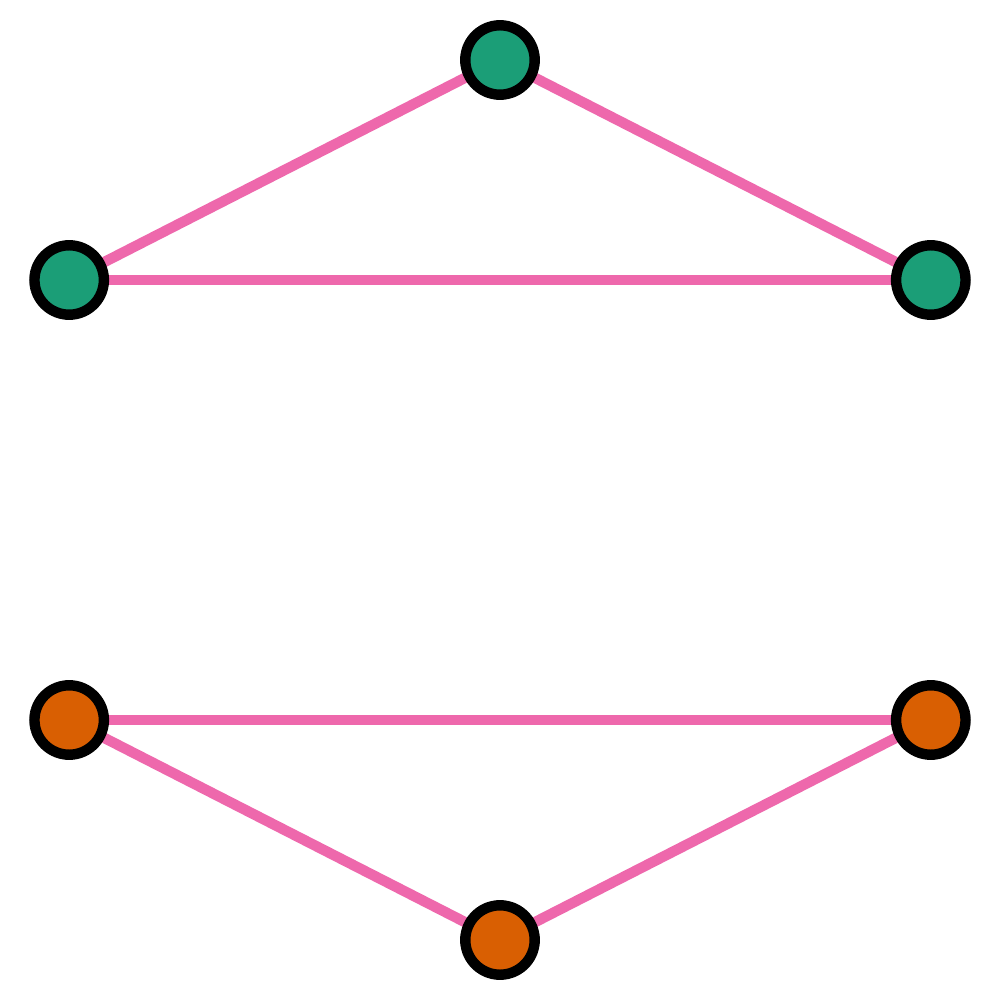}} & 
    \subfloat[\normalsize $h=0$, \ $\hat h = 0$]{\includegraphics[width=.20\textwidth]{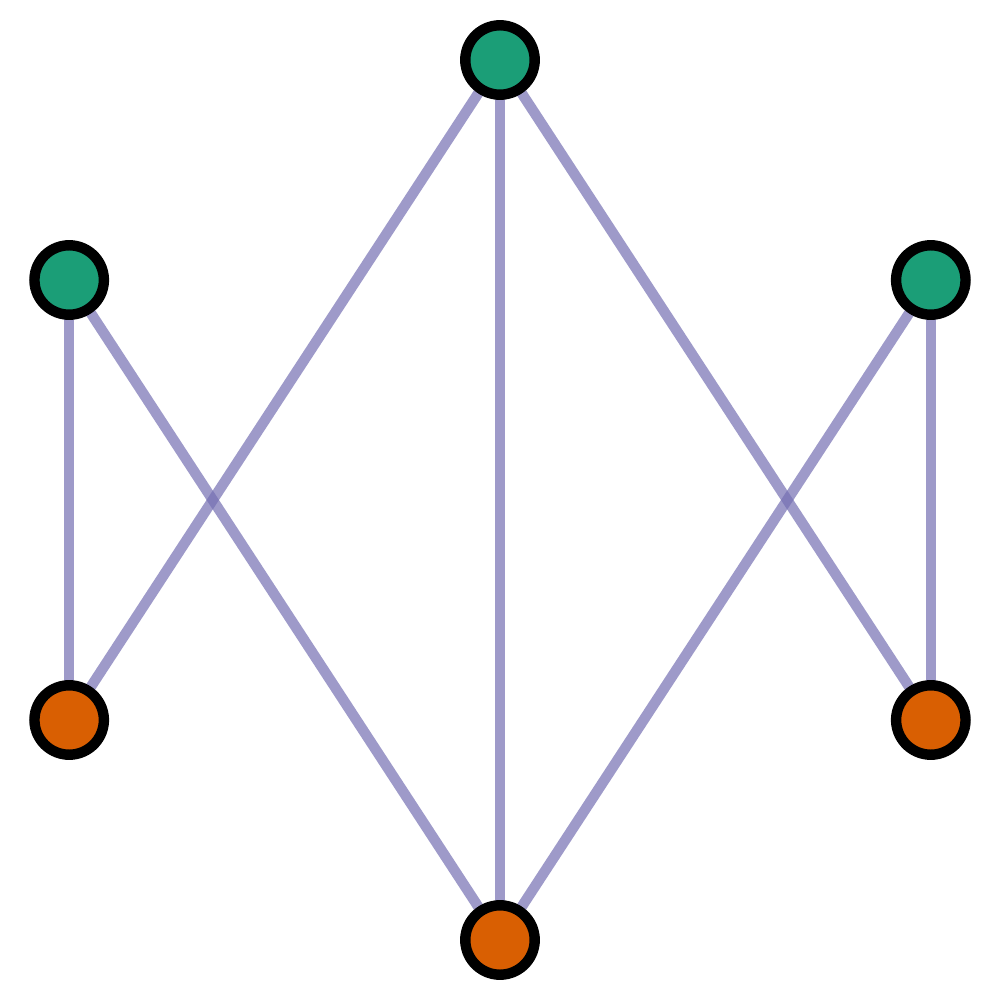}}\\
    \subfloat[\normalsize $h=.5$, \ $\hat h = 0$]{\includegraphics[width=.20\textwidth]{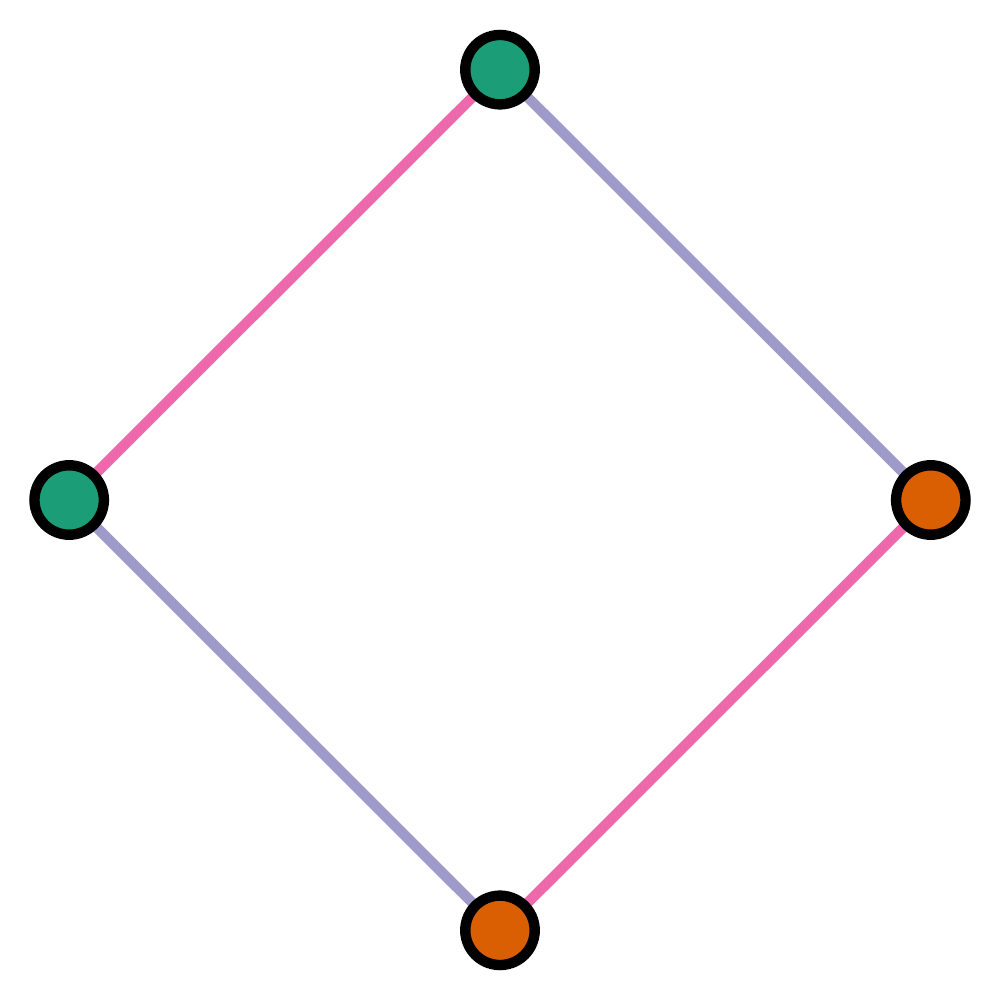}} & 
    \subfloat[\normalsize $h=.33$, \ $\hat h = 0$]{\includegraphics[width=.20\textwidth]{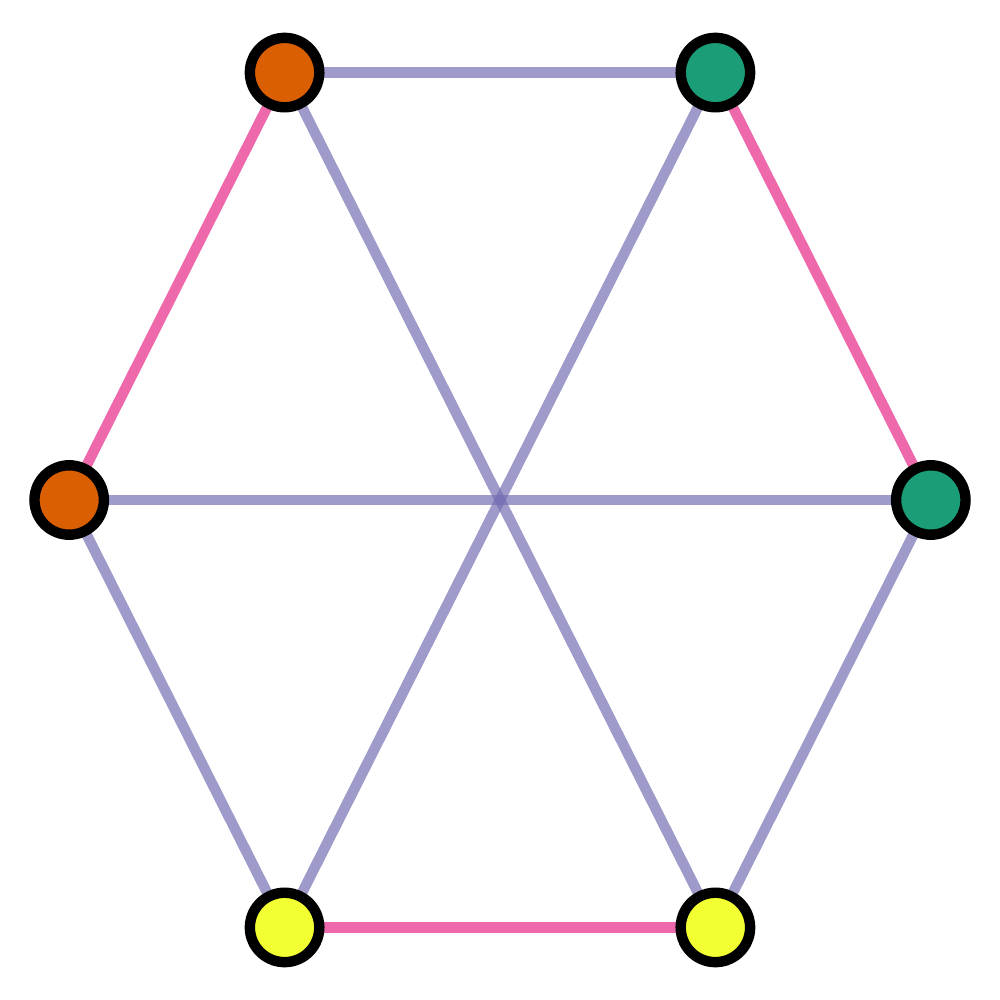}} \\
    \subfloat[\normalsize $h=.53$, \ $\hat h = .07$]{\includegraphics[width=.20\textwidth]{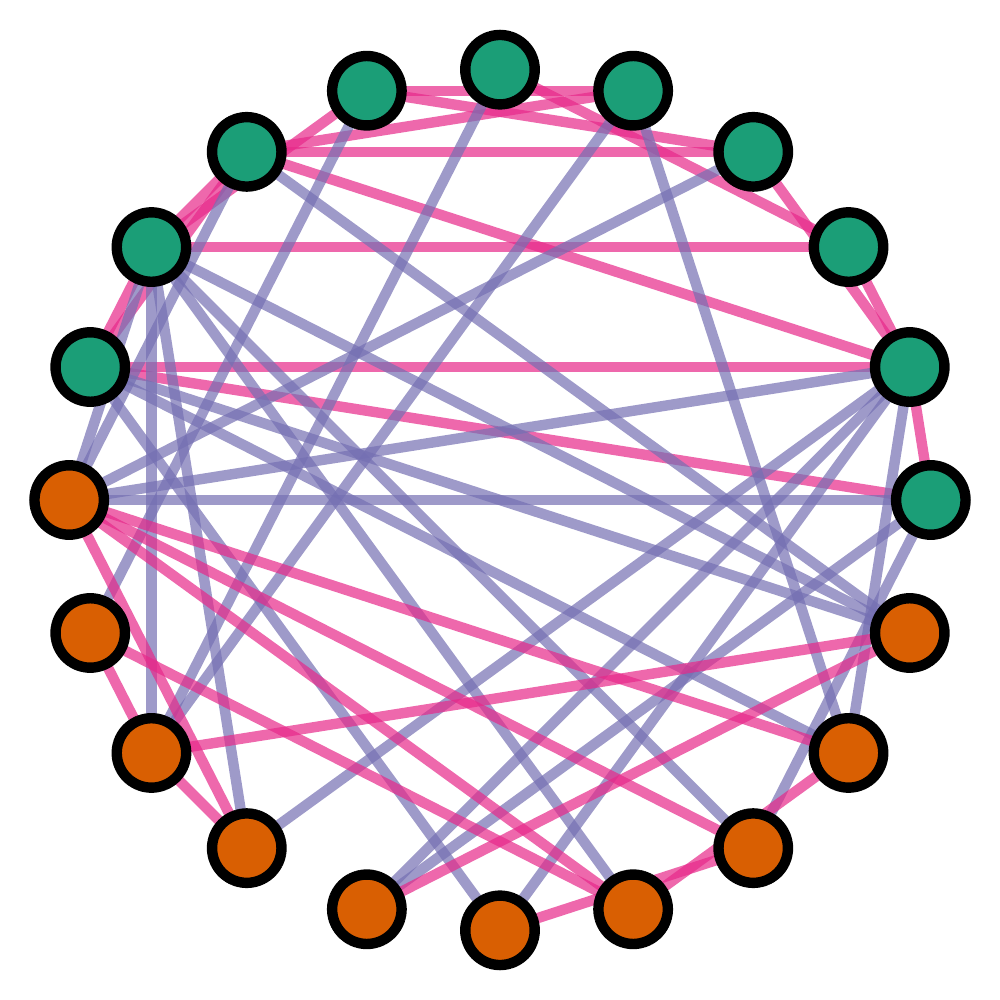}} &
    \subfloat[\normalsize $h=.66$, \ $\hat h = .04$]{\includegraphics[width=.20\textwidth]{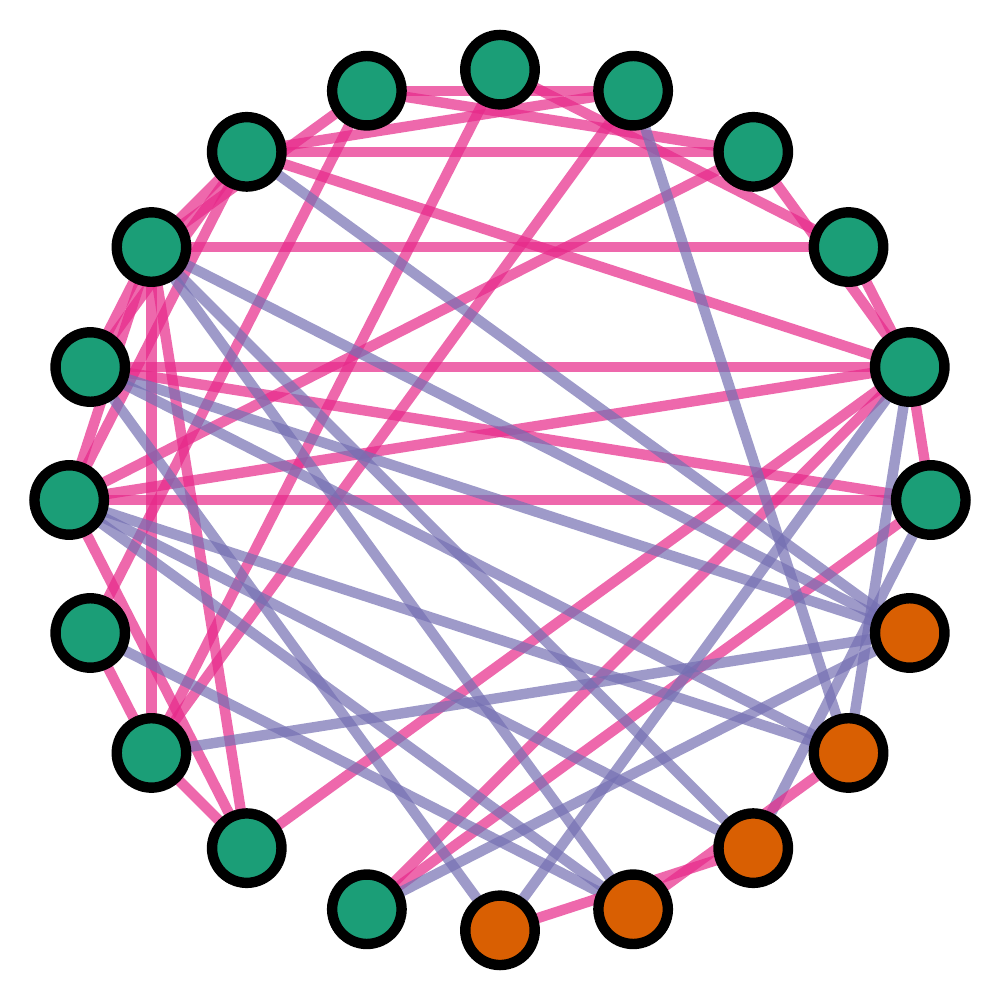}} 
    \end{tabular}
    \caption{Examples of graphs with different label-topology relationships and comparison of our measure $\hat h$ with the edge homophily ratio $h$. The node classes are labeled by color.  \textcolor[HTML]{e7298a}{Pink} edges link nodes of the same class, while \textcolor[HTML]{7570b3}{purple} edges link nodes of different classes. (a,b) Pure homophily and pure heterophily. Both measures equal $1$ in homophily and $0$ in heterophily.  (c,d) Graphs where each node is connected to one member of each class. Edge homophily depends on the number of classes, while our measure $\hat h$ does not. (e,f) Random Erd\H{o}s R\'enyi graphs in which edges are independent of labels. Edge homophily is sensitive to class imbalance, while our measure $\hat h$ is not.}
    \label{fig:example_graphs}
\end{figure}

\subsection{Proposed Datasets}

Here, we detail the non-homophilous datasets that we propose for graph machine learning evaluation. These datasets come from a variety of contexts, and some have been used for evaluation of graph machine learning models in past work; in certain cases, we make adjustments such as modifying node labels and adding node features. We define node features for Facebook100, Pokec, and snap-patents, while we redefine node labels for arXiv-year and snap-patents.
Basic dataset statistics are given in Table \ref{tab:data_stats}. The datasets are as follows:

\myparagraph{Twitch-explicit} contains 7 networks where Twitch users are nodes, and mutual friendships between them are edges \cite{rozemberczki2019multi}. Node features are games liked, location and streaming habits. Each graph is associated to users of a particular region. The class labels denote whether a streamer uses explicit language. We solely train and test on Twitch-DE, which has 9,498 nodes, 76,569 edges, edge homophily of .632, and $\hat h$  of .142.
    
\myparagraph{YelpChi} \cite{mukherjee2013yelp} is a graph in which the nodes are reviews for hotels and restaurants in the Chicago area, and the class labels are fraudulent reviews and recommended reviews. The 32 node features are taken from \cite{rayana2015collective}. We take the topology from \cite{dou2020enhancing};  while there are different relations captured by edges, we treat them all as the same single edge type.

\myparagraph{deezer-europe} \cite{rozemberczki2020characteristic} is a social network of users on Deezer from European countries, where edges represent mutual follower relationships. The node features are based on artists liked by each user. Nodes are labeled with reported gender.

\myparagraph{Facebook100} \cite{traud2012social} consists of 100 Facebook friendship network snapshots from 2005, each of which has as nodes the Facebook users from a given American university. Each node is labeled with the reported gender of the user. The node features are major, second major/minor, dorm/house, year, and high school. We solely train and test on Penn94, which has 41,554 nodes, 1,362,229 edges, edge homophily of .470, and $\hat h$ of .046.

\myparagraph{Pokec} is the friendship graph of a Slovak online social network, where nodes are users and edges are directed friendship relations \cite{snapnets}. Nodes are labeled with reported gender. The node features are derived from profile information, such as geographical region, registration time, and age.

\myparagraph{ogbn-proteins} \cite{hu2020open} has proteins as nodes and different biological relationships between proteins as edges. There are 112 tasks, in which each protein is given a binary label. Also, there are no separate node features --- instead, the means of incoming edge features give 8 dimensional node features.

\myparagraph{arXiv-year} \cite{hu2020open} is the ogbn-arxiv network with class labels given by the year that the paper is posted, instead of by subject areas. The nodes are arXiv papers, and directed edges connect a paper to other papers that it cites. The node features are averaged word2vec token features of both the title and abstract of the paper. The five classes are chosen by partitioning the posting dates so that class ratios are approximately balanced: 2013 and prior, 2014-2015, 2016-2017, 2018, and 2019-2020.

\myparagraph{snap-patents} \cite{leskovec2005graphs, snapnets} is a dataset of utility patents granted between 1963 to 1999 in the US. Each node is a patent, and edges connect patents that cite each other. Node features are derived from patent metadata. The task is to predict the time at which a patent was granted. The five classes are:  1971 and prior, 1972-1980, 1981-1988, 1989-1994, and 1995-1999.

\subsection{General Non-homophilous Settings}

Different settings in which non-homophilous relationships are prevalent have been identified in the literature and are represented by our proposed datasets:
\begin{itemize}
    \item Gender relations in social or interaction networks  \cite{altenburger2018monophily, chin2019decoupled, jia2020residual} (deezer, FB100, Pokec).
    \item Biological structures such as in food webs \cite{gatterbauer2014semi} and protein interactions \cite{newman2003mixing} (ogbn-proteins).
    \item Technological and internet relationships, such as in web page connections \cite{newman2003mixing, pei2019geom}.
    \item Malicious or fraudulent nodes, such as in auction networks \cite{chau2006detecting, pandit2007netprobe} (YelpChi).
    \item Publication time in citation networks \cite{peel2017graph} (arXiv-year, snap-patents).
\end{itemize}
While not all example graph data from these contexts are non-homophilous, a diverse range are. In order to succeed in future applications in these contexts, it may be important to develop methods that are able to handle non-homophilous structures.

\section{Experiments}

\begin{table*}[ht]
    \centering
    \caption{Experimental results. Test accuracy is displayed for most datasets, while Twitch-DE, YelpChi, and ogbn-proteins display test ROC AUC$\dagger$. Standard deviations are over 5 train/val/test splits, except for ogbn-proteins, which has a fixed split. The three best results per dataset are highlighted in \colorbox{blue!25}{blue}. (M) denotes some  (or all) hyperparameter settings run out of memory.}
    \label{tab:results}
    {\footnotesize
    \begin{tabular}{cllllllll}
    \toprule
     & Twitch-DE$\dagger$ & YelpChi$\dagger$ & deezer & Penn94 (FB100)  & pokec &  ogbn-proteins$\dagger$ & arXiv-year & snap-patents  \\
    \midrule
     MLP & $69.20\std{0.62}$ & \bestcell $87.94\std{0.52}$ & $66.55\std{0.72}$ & $73.61\std{0.40}$ & $62.37\std{0.02}$ & $73.43 \std{0.12}$ & $36.70\std{0.21}$ & $31.34\std{0.05}$  \\    
     \hdashline
     L Prop (1 hop) & $71.00\std{1.01}$  & $63.64\std{0.73}$  & $56.50\std{0.41}$ & $63.21\std{0.39}$ & $53.09\std{0.05}$ & \bestcell $75.14\std{0.00}$ & $43.42\std{0.17}$ & $30.28\std{0.09}$   \\    
     L Prop (2 hop) & $72.27\std{0.78}$  & $62.34\std{0.78}$ & $56.96\std{0.26}$ & $74.13\std{0.46}$ & $76.76\std{0.03}$ & $65.79\std{0.00}$ & $46.07\std{0.15}$ & $38.61\std{0.07}$ \\
     LINK & $72.42\std{0.57}$ & $63.44\std{1.07}$  & $57.71\std{0.36}$ & $80.79\std{0.49}$ & \bestcell $80.54\std{0.03}$ & $63.49 \std{0.02}$ & \bestcell $53.97\std{0.18}$ & \bestcell $60.39\std{0.07}$ \\    
     \hdashline
     SGC (1 hop) & $72.30\std{0.22}$ & $58.62\std{0.85}$ & $59.73\std{0.12}$ & $66.79\std{0.27}$ & $53.61\std{0.17}$ & $50.58\std{1.08}$ &  $32.83\std{0.13}$ & $30.31\std{0.06}$   \\    
     SGC (2 hop) & \bestcell $73.65\std{0.40}$ & $57.18\std{0.75}$  & $61.56\std{0.51}$ & $76.09\std{0.45}$ & $62.81\std{1.42}$ & $49.48\std{1.45}$ & $32.27\std{0.06}$ & $29.09\std{0.09}$  \\    
     C\&S (1 hop) & $68.54\std{1.25}$ & $87.47\std{0.50}$ & $64.60\std{0.57}$ & $72.55\std{0.63}$ & $63.23\std{0.07}$ & $71.13\std{0.69}$ & $42.08\std{0.27}$ & $37.41\std{0.07}$ \\    
     C\&S (2 hop) & $69.39\std{0.85}$ & $87.40\std{0.44}$ & $64.52\std{0.62}$ & $72.47\std{0.73}$ &  $77.22\std{0.61}$ & $69.31\std{0.62}$ & $42.17\std{0.27}$ & $44.62\std{0.14}$ \\    
     \hdashline
     GCN & \bestcell $74.07 \std{0.68}$ & $63.62 \std{1.00}$ & $62.23\std{0.53}$ & \bestcell $82.47\std{0.27}$ & $75.45\std{0.17}$ & $72.03 \std{0.32}$ & $46.02\std{0.26}$ & $45.65\std{0.04}$ \\
     GAT & $73.13\std{0.29}$ & $81.42 \std{2.12}$ (M) & $61.09\std{0.77}$ & $81.53\std{0.55}$ & $71.77\std{6.18}$ (M) & (M) & $46.05\std{0.51}$ & $45.37\std{0.44}$ (M)  \\
     GCN+JK & $70.57 \std{2.22}$ & $64.35 \std{0.86}$	& $60.99\std{0.14}$ & $81.63\std{0.54}$ & $77.00\std{0.14}$ & \bestcell $76.88 \std{0.48}$ &  $46.28\std{0.29}$ & \bestcell $46.88\std{0.13}$  \\
     GAT+JK & $73.06 \std{0.49}$  & \bestcell  $90.04 \std{0.61}$ (M) & $59.66\std{0.92}$ & $80.69\std{0.36}$ & $71.19\std{6.96}$ (M) & (M) & $45.80\std{0.72}$ & $44.78\std{0.50}$ (M) \\
     APPNP & $72.20\std{0.73}$ & $86.66\std{0.64}$ & \bestcell $67.21\std{0.56}$ & $74.95\std{0.45}$ & $62.58\std{0.08}$ & $68.96\std{0.09}$ & $38.15\std{0.26}$  & $32.19\std{0.07}$ \\
     \hdashline
     H\textsubscript{2}GCN & $72.67\std{0.65}$ & \bestcell $88.48\std{0.21}$  & \bestcell $67.22\std{0.90}$ & (M) & (M) & (M) & \bestcell $49.09\std{0.10}$ & (M) \\
     MixHop & $73.23\std{0.99}$ & $87.02\std{0.50}$  & $66.80\std{0.58}$ & \bestcell $83.47\std{0.71}$ & \bestcell $81.07\std{0.16}$ & \bestcell $75.60\std{0.85}$ & \bestcell $51.81\std{0.17}$ & \bestcell $52.16\std{0.09}$ (M) \\
     GPR-GNN & \bestcell $73.84\std{0.69}$ & $86.57\std{0.89}$ & \bestcell $66.90\std{0.50}$ & \bestcell $84.59\std{0.29}$ & \bestcell $78.83\std{0.05}$ & (M)  & $45.07\std{0.21}$  & $40.20\std{0.03}$   \\
    \bottomrule
    \end{tabular}
    }
    \vspace{10pt}
\end{table*}

\subsection{Experimental Setup}

We include both methods that are graph-agnostic and node-feature-agnostic as simple baselines; the node-feature-agnostic models of two-hop label propagation \cite{peel2017graph} and LINK (logistic regression on the adjacency matrix) \cite{zheleva2009to} have been found to perform well in various non-homophilous settings, but they have often been overlooked by recent graph representation learning work. Also, we include SGC \cite{wu2019simplifying} and C\&S \cite{huang2021combining} as simple methods that perform well on homophilic datasets. We include a two-hop propagation variant of C\&S in analogy with two-step label propagation. In addition to representative general GNNs, we also include three GNNs recently proposed for non-homophilous settings. The full list of methods is:
\begin{itemize}
\setlength\itemsep{3pt}
    \item Models that only use node features: MLP \cite{goodfellow2016deep}.
    \item Models that only use the graph topology: label propagation (standard and two-hop) \cite{zhou2004learning, peel2017graph}, LINK \cite{zheleva2009to}.
    \item Simple methods: SGC \cite{wu2019simplifying}, C\&S \cite{huang2021combining}, two-hop variants.
    \item General GNNs: GCN \cite{kipf2017semi}, GAT \cite{velivckovic2018graph}, jumping knowledge networks (GCN+JK, GAT+JK) \cite{xu2018representation}, and APPNP \cite{klicpera2019predict}.
    \item Non-homophilous GNNs: H\textsubscript{2}GCN \cite{zhu2020beyond}, MixHop \cite{abu2019mixhop}, and GPR-GNN \cite{chien2021adaptive}.
\end{itemize}

Following other works in non-homophilous graph learning evaluation, we take a high proportion of training nodes \cite{zhu2020beyond, pei2019geom, yan2021two}; we run each method on the same five random 50/25/25 train/val/test splits for each dataset, besides ogbn-proteins, for which we use the original Open Graph Benchmark splits \cite{hu2020open}. All methods requiring gradient-based optimization are run for 500 epochs, with test performance reported for the learned parameters of highest validation performance. We use ROC-AUC as the metric for the class-imbalanced Twitch-DE (60.5\% of nodes in majority class), YelpChi (85.5\% of nodes in majority class), and ogbn-proteins datasets (52.7\% to 98.0\% in majority, depending on task). For other datasets, we use classification accuracy as the metric. Further experimental details can be found in Appendix \ref{sec:exp_details}.

\subsection{Experimental Results}

Table \ref{tab:results} lists the results of each method across the datasets that we propose. Our new measure and new datasets reveal several important properties of non-homophilous node classification. Firstly, both methods that only use node features and methods that only use graph topology appear to perform better than random, thus demonstrating the quality of our datasets. Secondly, the stability of performance across runs is better for our datasets than those of \citet{pei2019geom} (see \cite{zhu2020beyond} results). Moreover, as suggested by prior theory and experiments \cite{zhu2020beyond, abu2019mixhop, chien2021adaptive}, the non-homophilous GNNs usually do well --- though not necessarily on every dataset.

The core assumption of homophily in SGC and C\&S that enables them to be simple and efficient does not hold on these non-homophilous datasets, and thus the performance of these methods is typically relatively low. Nevertheless, 2-hop C\&S still achieves high performance on some datasets. Indeed, our results indicate that simple 2-hop modifications of learning methods often improve performance in low-homophily, though we note that 1-hop label propagation performs much better than expected on ogbn-proteins, possibly due to some implementation nuance. LINK, a frequently ignored baseline that in some sense acts on two-hop neighborhoods of each node \cite{altenburger2018monophily}, performs well on many datasets --- despite not using node feature information.

Finally, one consequence of using larger datasets for benchmarks is that the tradeoff between scalability and learning performance of non-homophilous methods has become starker, with some methods facing memory issues. This tradeoff is especially important to consider in light of the fact that many scalable graph learning methods rely on implicit or explicit homophily assumptions \cite{wu2019simplifying, huang2021combining, deng2020graphzoom, bojchevski2020scaling}, and thus face issues when used in non-homophilous settings.

\section{Conclusion}

In this paper, we introduce a measure of the presence of homophily that alleviates issues with existing measures, propose new, high-quality non-homophilous graph learning datasets, and benchmark simple baselines and representative graph representation learning methods across our datasets. We hope that these contributions will provide researchers of non-homophilous graph learning methods with better tools to test their models and evaluate the utility of new techniques. 
While we benchmark on transductive node classification, the datasets we propose could be adapted to benchmark link prediction, clustering tasks, and inductive node classification in the case of Twitch-explicit and Facebook100. Future work could study these other tasks in low-homophily settings, reformulate current understandings of homophily that are most natural in node classification, and introduce new benchmarks for a wider range of applications.

\clearpage

\begin{acks}
 We thank Austin Benson and Horace He for insightful discussions. We also thank the rest of Cornell University Artificial Intelligence for their support and discussion. This research was supported by Facebook AI.
\end{acks}

\bibliography{refs}
\bibliographystyle{ACM-Reference-Format}

\appendix

\section{Compatibility Matrices and Statistics}\label{sec:appendix_measures}

\begin{figure*}
\centering
\begin{tabular}{ccc}
    \includegraphics[width=.26\textwidth]{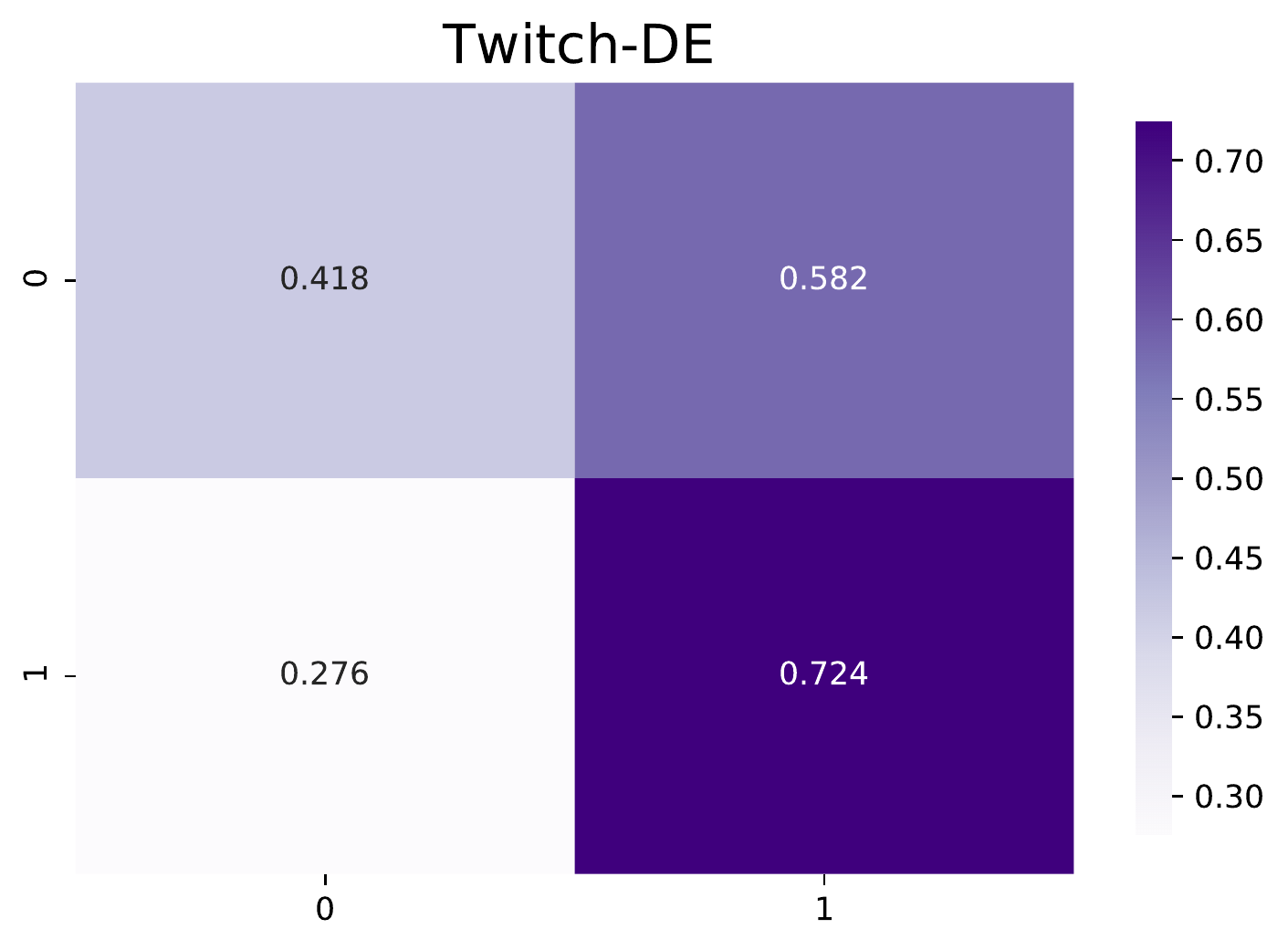} &
    \includegraphics[width=.26\textwidth]{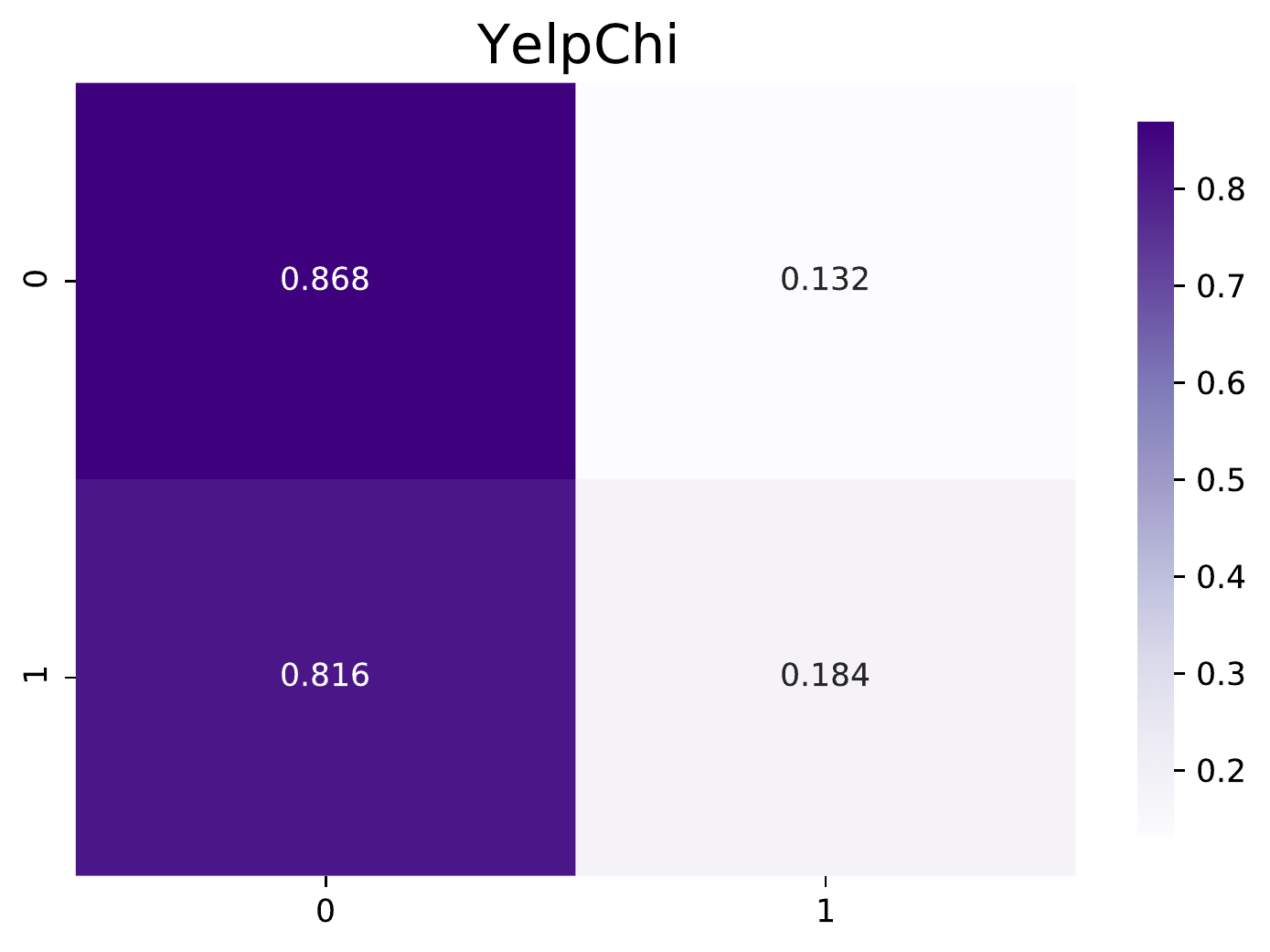} &
    \includegraphics[width=.26\textwidth]{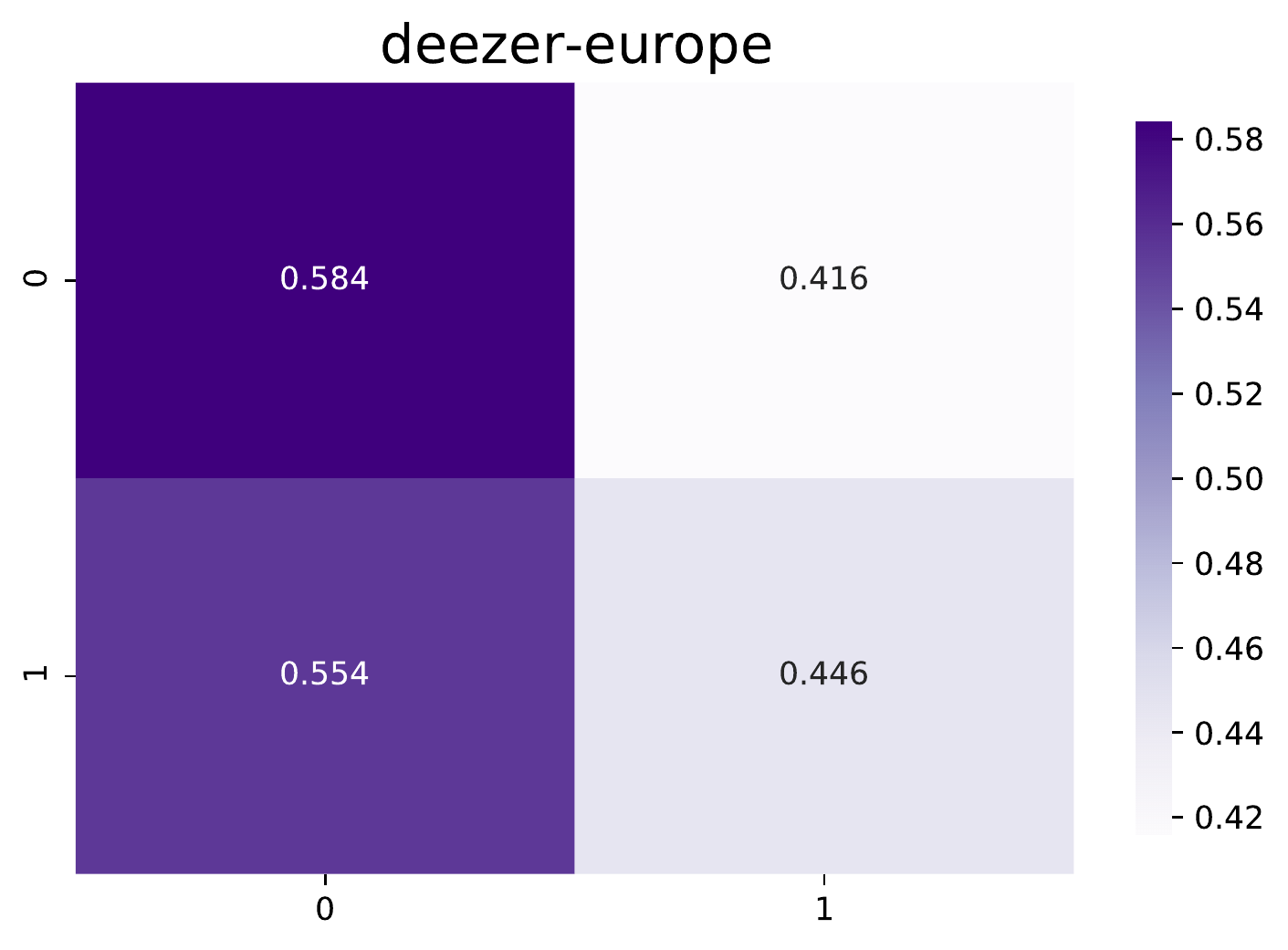} \\
     \includegraphics[width=.26\textwidth]{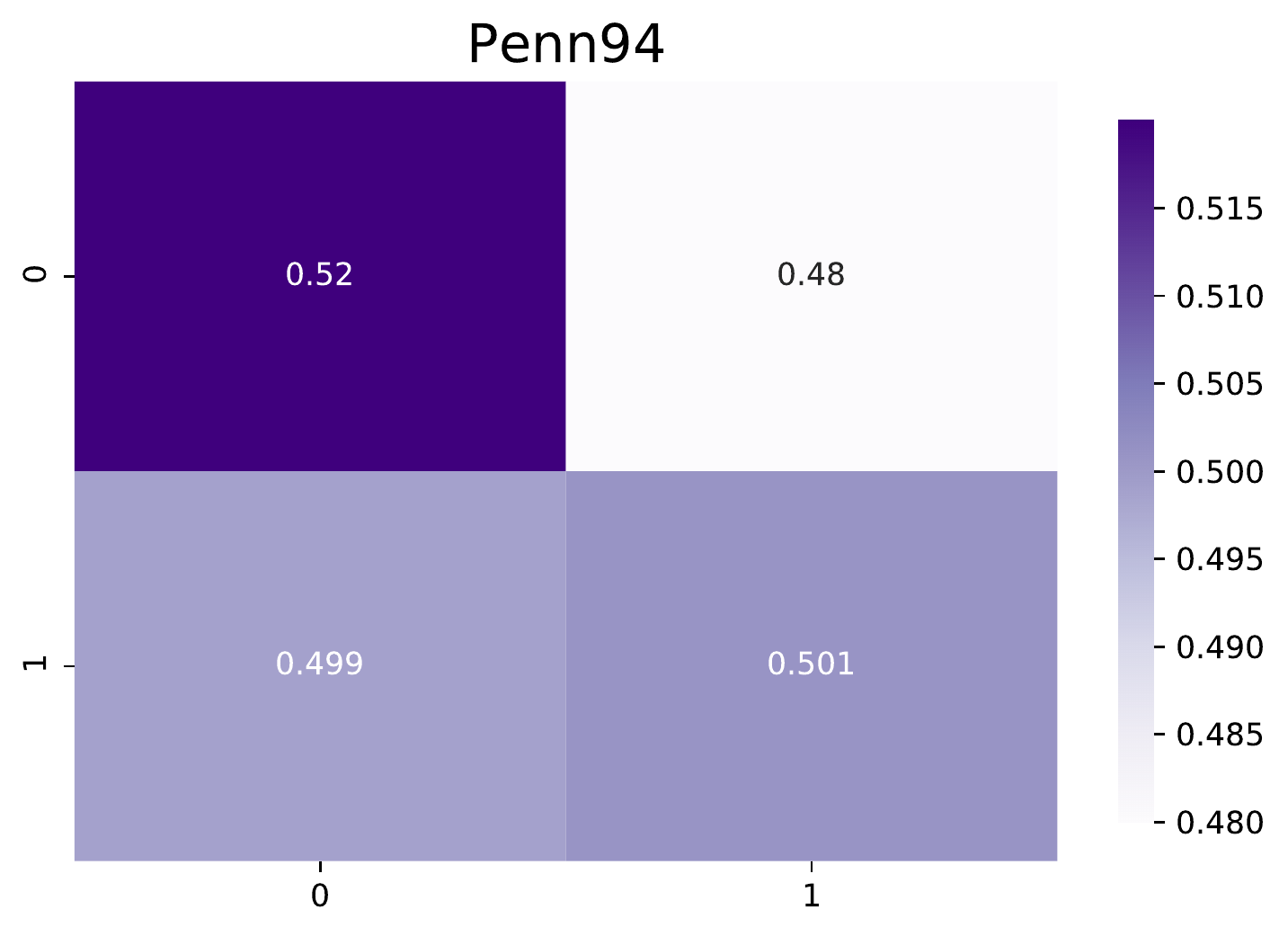} & 
    \includegraphics[width=.26\textwidth]{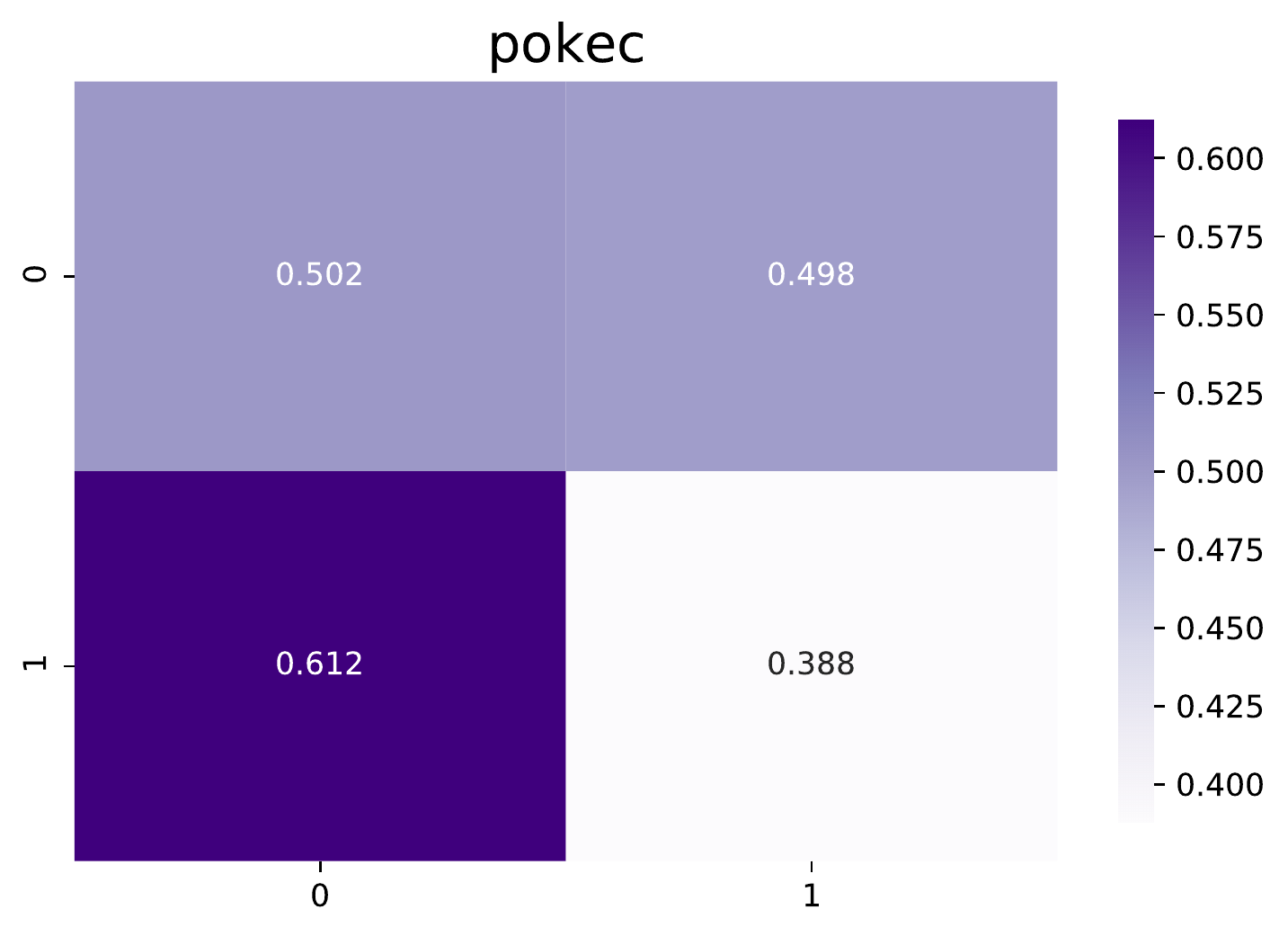} &
    \includegraphics[width=.26\textwidth]{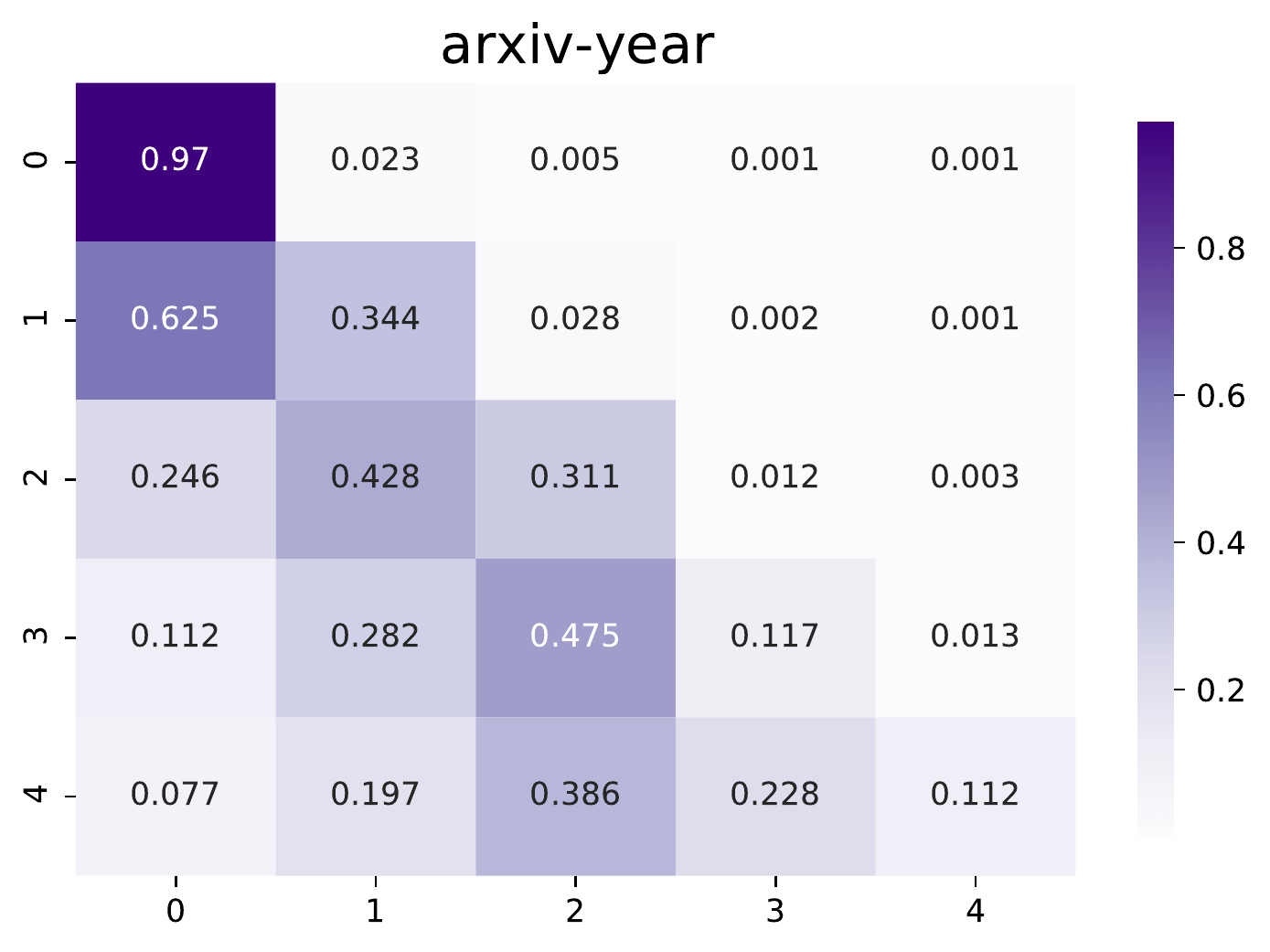} \\
    \includegraphics[width=.26\textwidth]{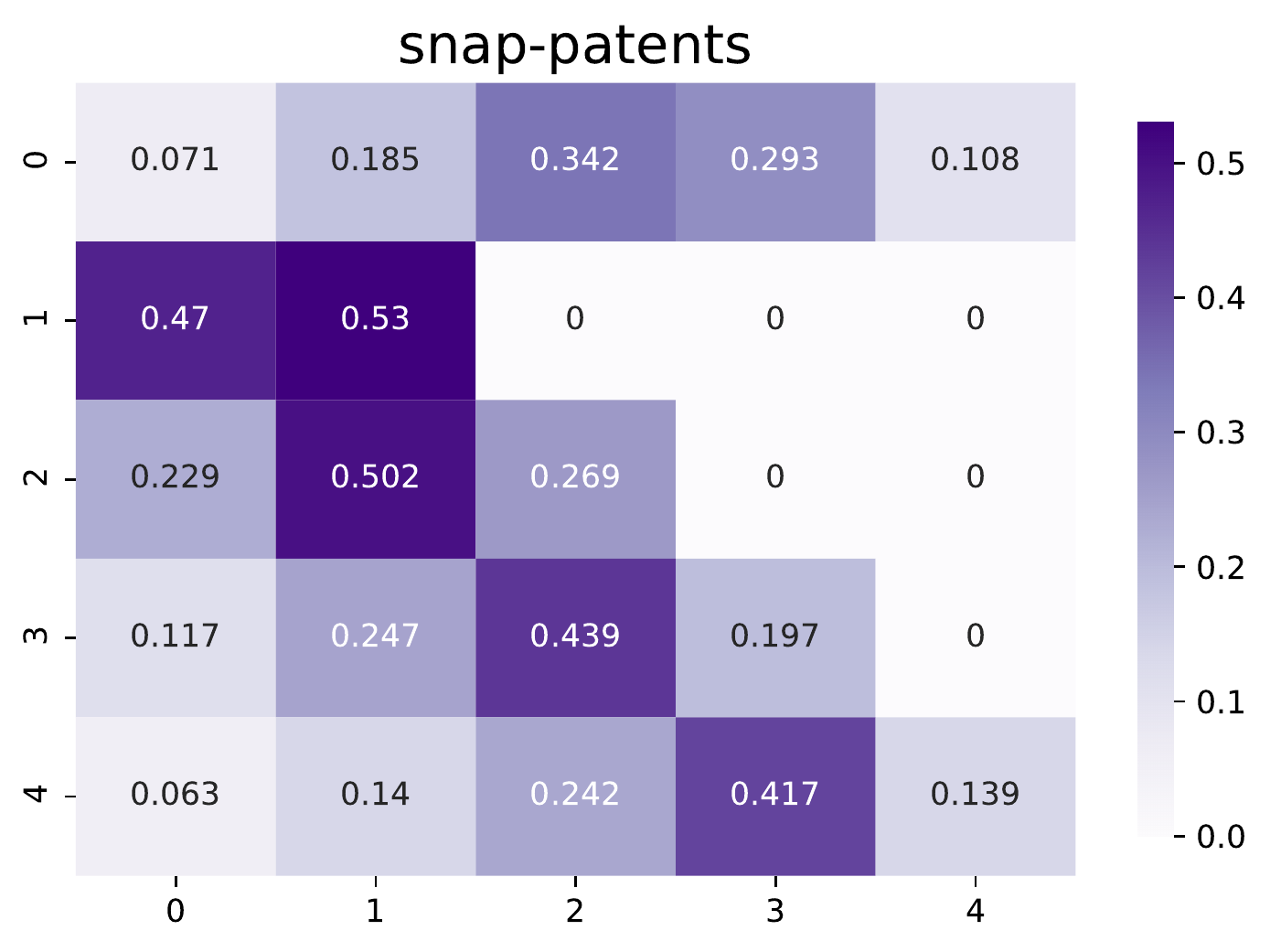} &
    \includegraphics[width=.26\textwidth]{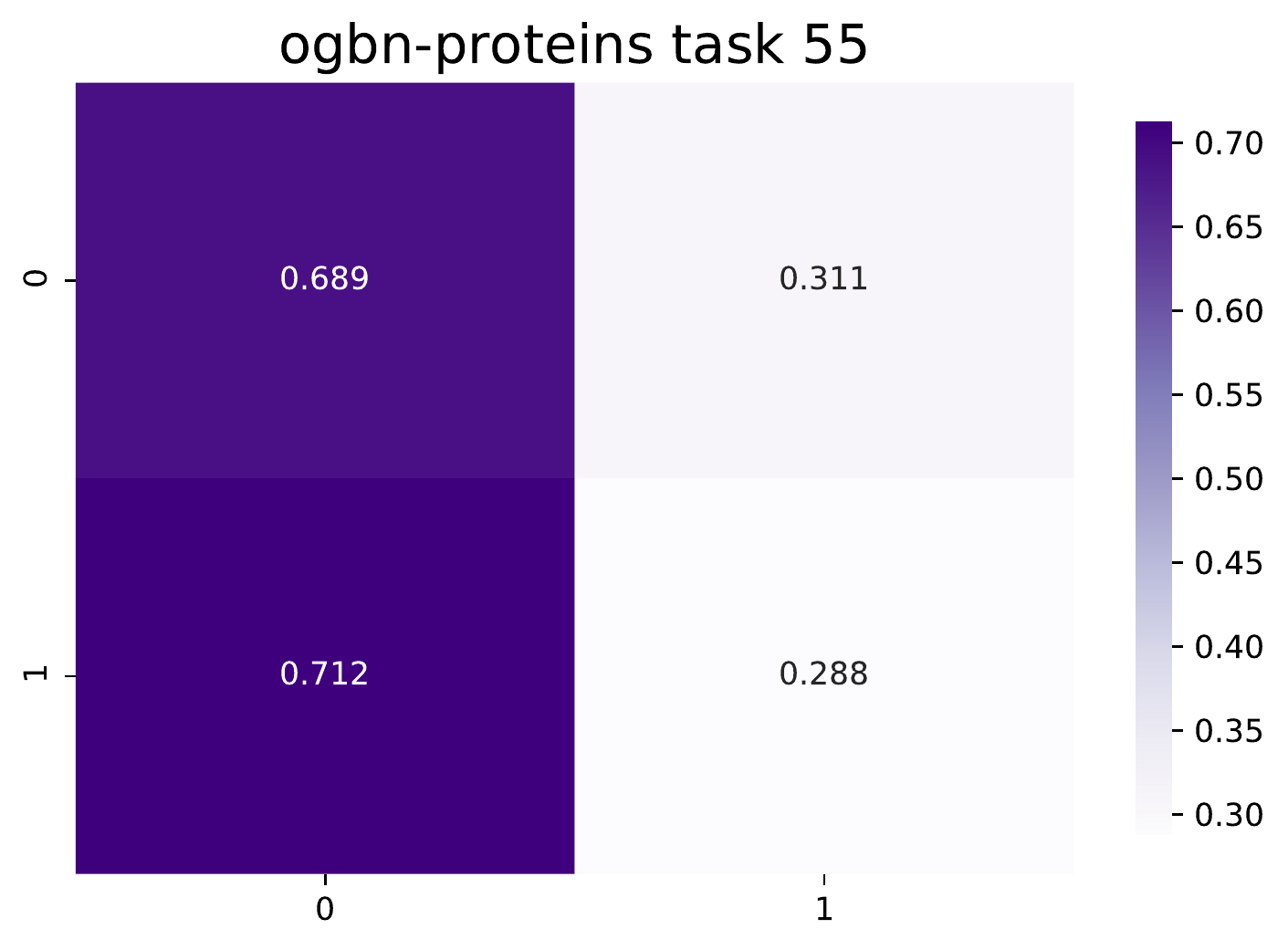} &
    \includegraphics[width=.26\textwidth]{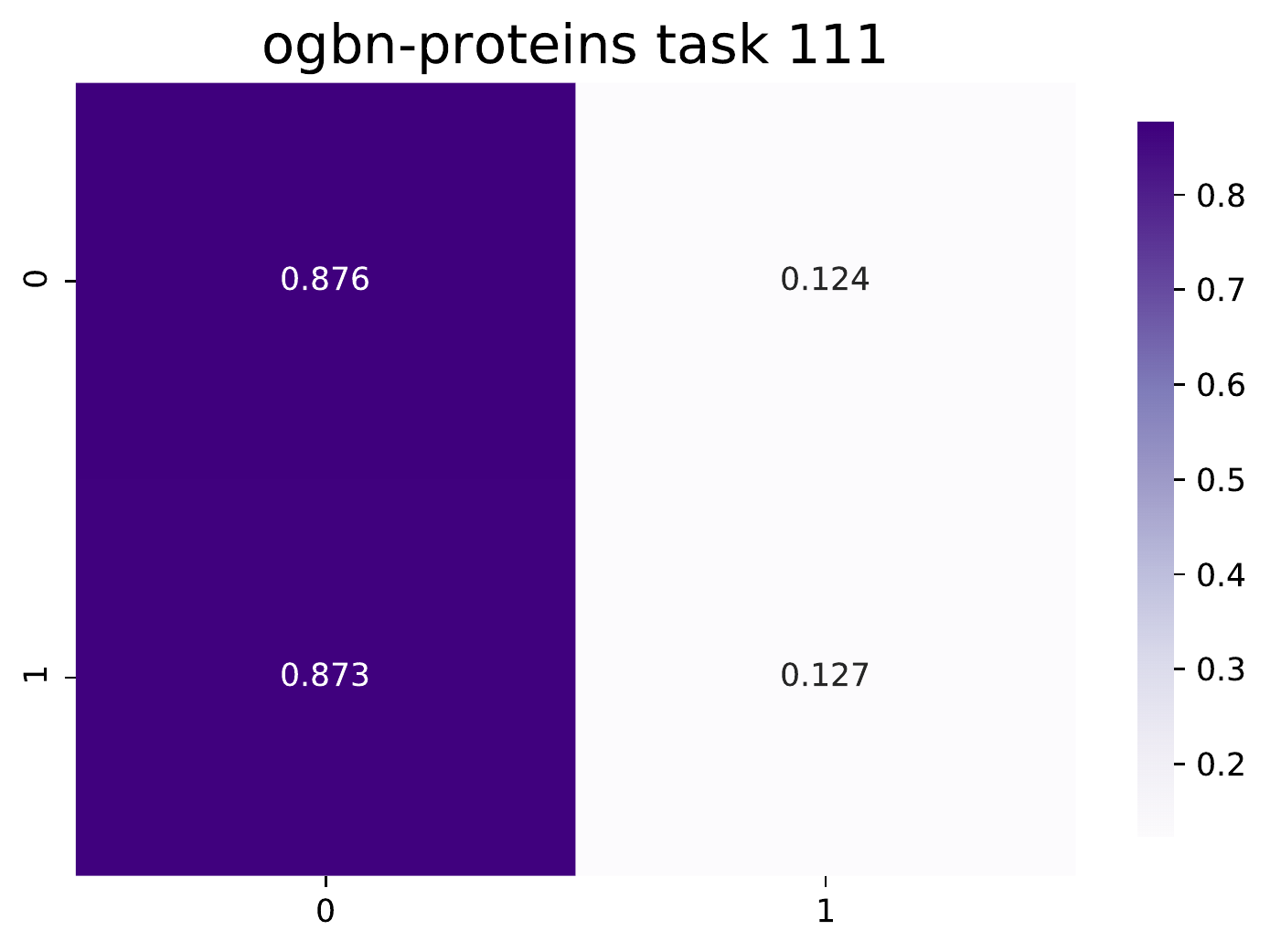} 
\end{tabular}
\caption{Compatibility matrices of our proposed datasets. These datasets from a variety of different contexts exhibit a wide range of non-homophilous structures. We choose 2 of the 112 ogbn-proteins tasks to display.}
\label{fig:compat_ours}
\end{figure*}

Following previous work \cite{zhu2020beyond}, for a graph $G$ with $C$ node classes we define the $C \times C$ compatibility matrix $\mbf H$ by
\begin{equation}
    \mbf H_{kl} = \frac{|(u,v) \in E : k_u =k, \; k_v = l| }{|(u,v) \in E : k_u = k| }
\end{equation}

This captures finer details of label-topology relationships in graphs than single scalar metrics (like edge homophily and our $\hat h$) capture. For classes $k$ and $l$, the entry $\mbf H_{kl}$ measures the proportion of edges from nodes of class $k$ that are connected to nodes of class $l$.
Compatibility matrices for our proposed datasets are shown in Figure \ref{fig:compat_ours}. As evidenced by the different patterns, the proposed datasets show interesting types of label-topology relationships besides homophily. For instance, the citation datasets arXiv-year and snap-patents have primarily lower-triangular structure, since most citations reference past work. The Penn94 matrix is mostly uniform, so there is little to no gender preference in aggregate, though past work has shown other useful signals in the social structure of gender \cite{altenburger2018monophily}. In Twitch-DE, while streamers that use explicit content (class 1) often connect to other streamers of class 1, streamers that do not use explicit content (class 0) also often connect to streamers of class 1, thus giving an overall non-homophilous structure. In Pokec, there is some heterophily, in that one gender has some preference for friends of another gender.

\subsection{Homophilous Data Statistics}\label{sec:homophilous_stats}

\begin{table}[h]
    \centering
    \caption{Statistics for homophilic graph datasets. \# C is the number of node classes.}
    \label{tab:homophilic_stats}
    {\footnotesize
    \begin{tabular}{crrrrrr}
    \toprule
    Dataset & \# Nodes & \# Edges &   \# C &  Edge hom. & $\hat h$ (ours) \\
    \midrule
         Cora & 2,708 & 5,278 &  7 &  .81 & .766\\
         Citeseer & 3,327 & 4,552 &  6 &  .74 & .627\\
         Pubmed & 19,717 & 44,324 &  3 &  .80 & .664\\
         ogbn-arXiv & 169,343 & 1,166,243 &  40 &  .66 & .416\\
         ogbn-products & 2,449,029 & 61,859,140 &  47 &  .81 & .459 \\
         oeis & 226,282 & 761,687  & 5 &  .50 & .532 \\
    \bottomrule
    \end{tabular}
    }
\end{table}

\begin{figure*}[t]
\centering
\begin{tabular}{ccc}
    \includegraphics[width=.27\textwidth]{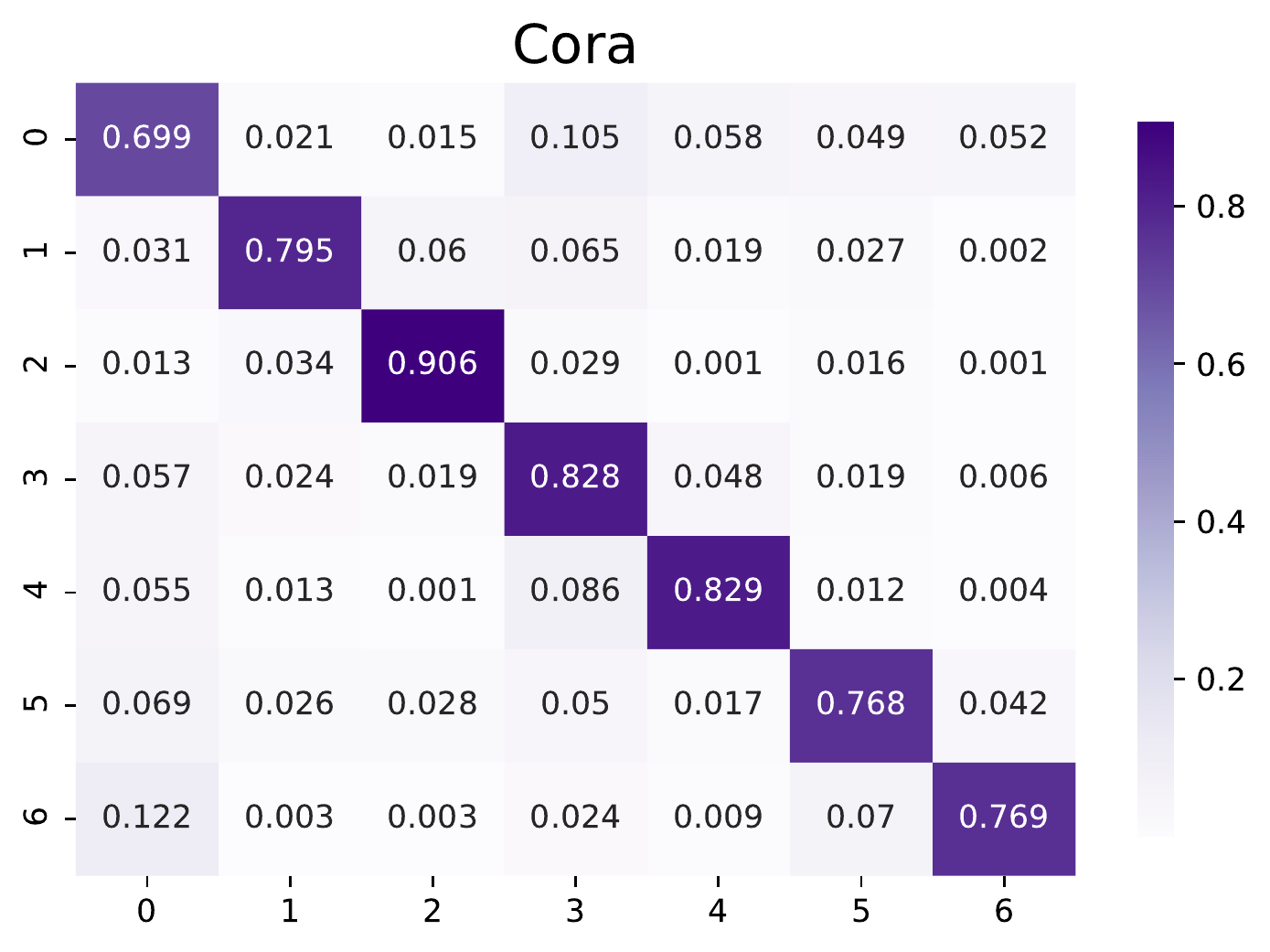} &
    \includegraphics[width=.27\textwidth]{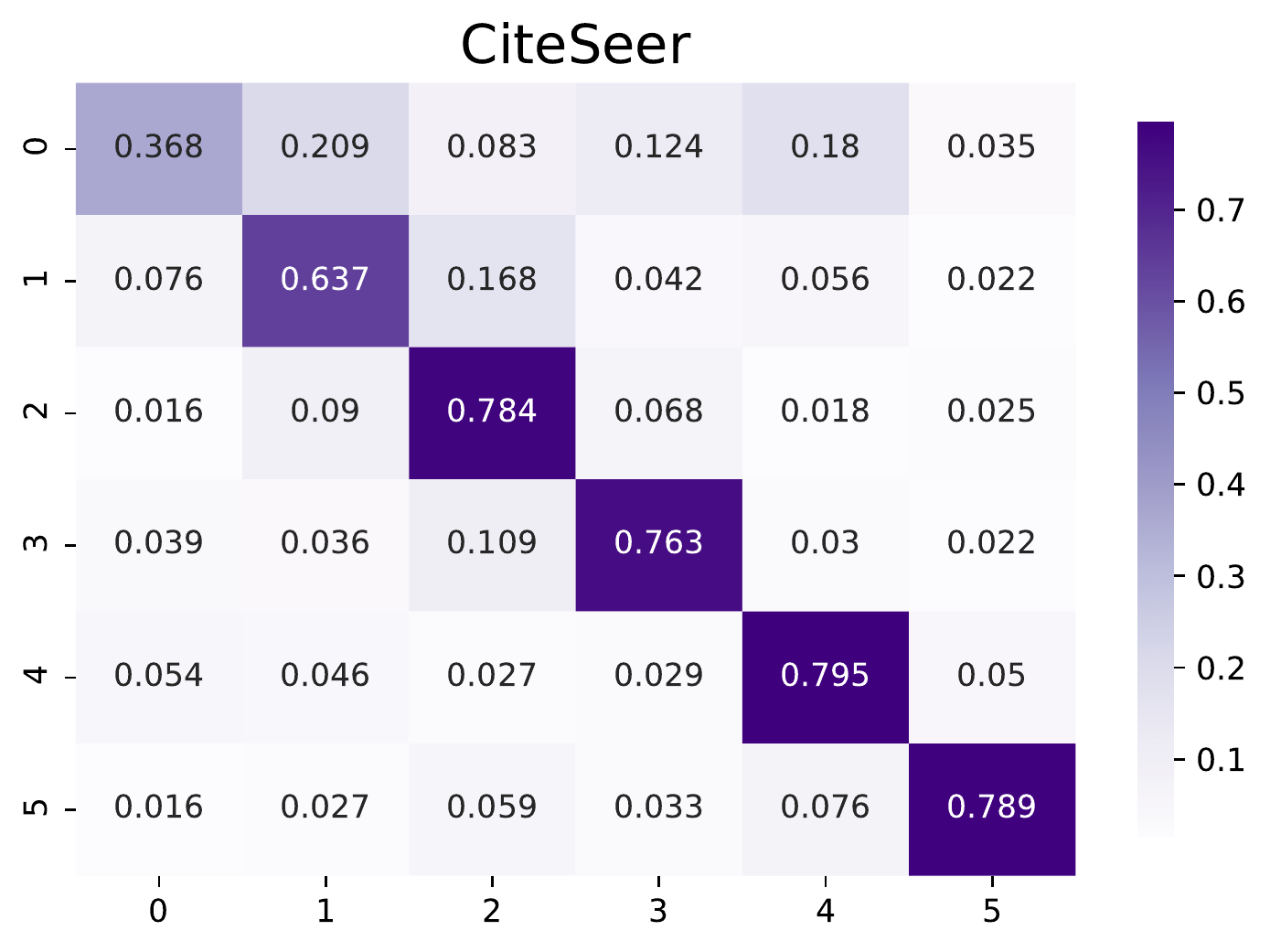} &
    \includegraphics[width=.27\textwidth]{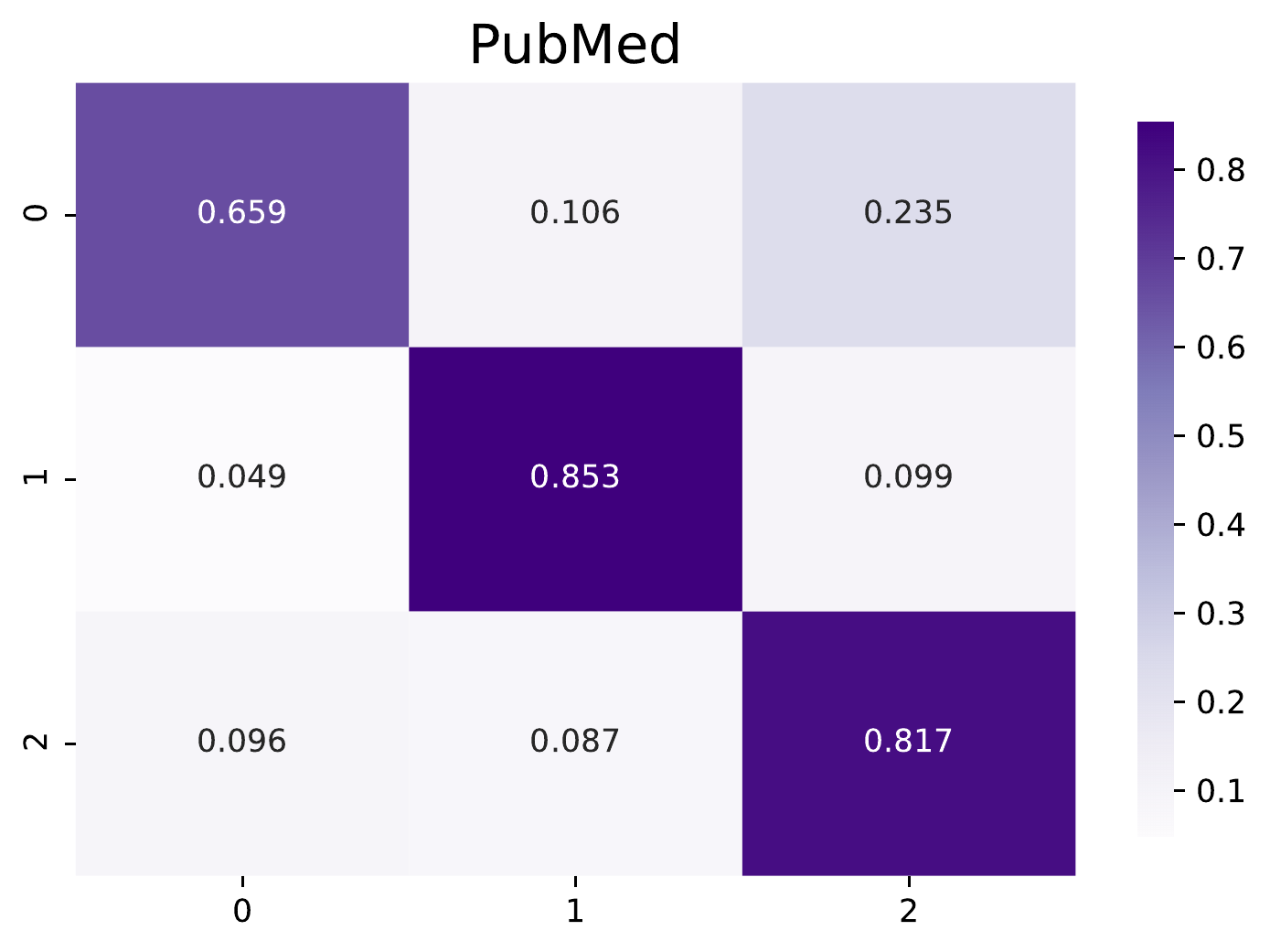} \\
    \includegraphics[width=.27\textwidth]{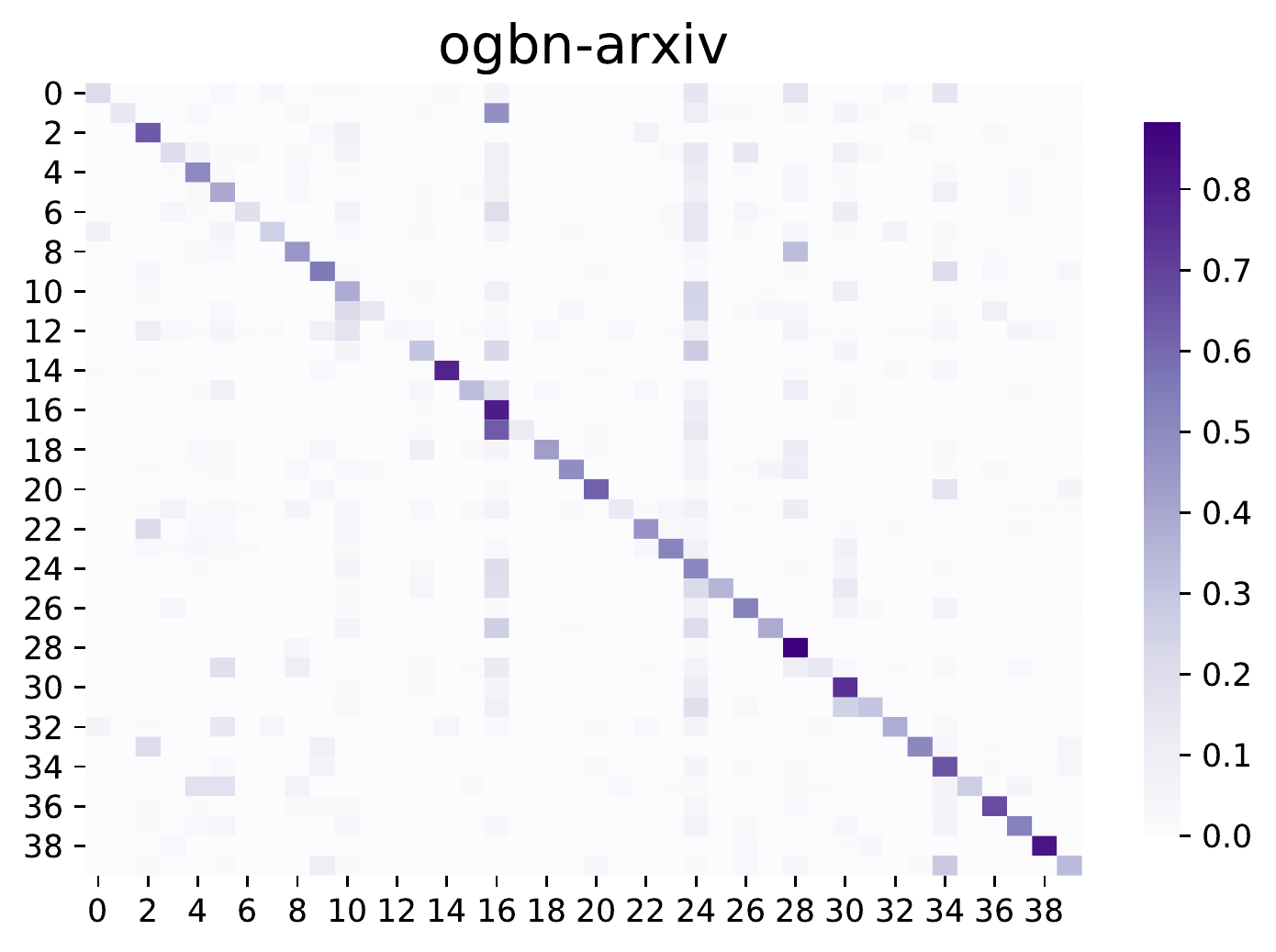} &
    \includegraphics[width=.27\textwidth]{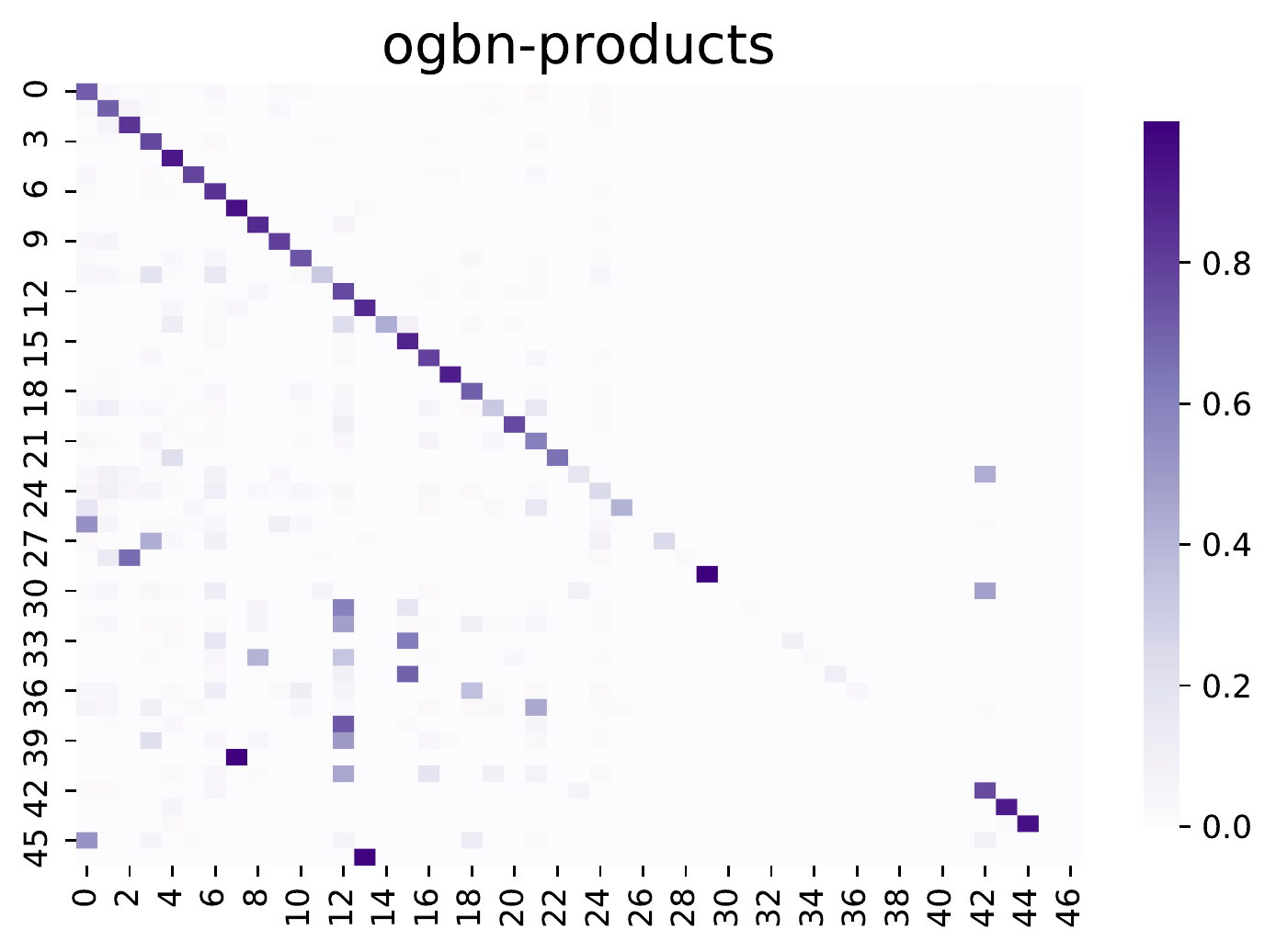} &
    \includegraphics[width=.27\textwidth]{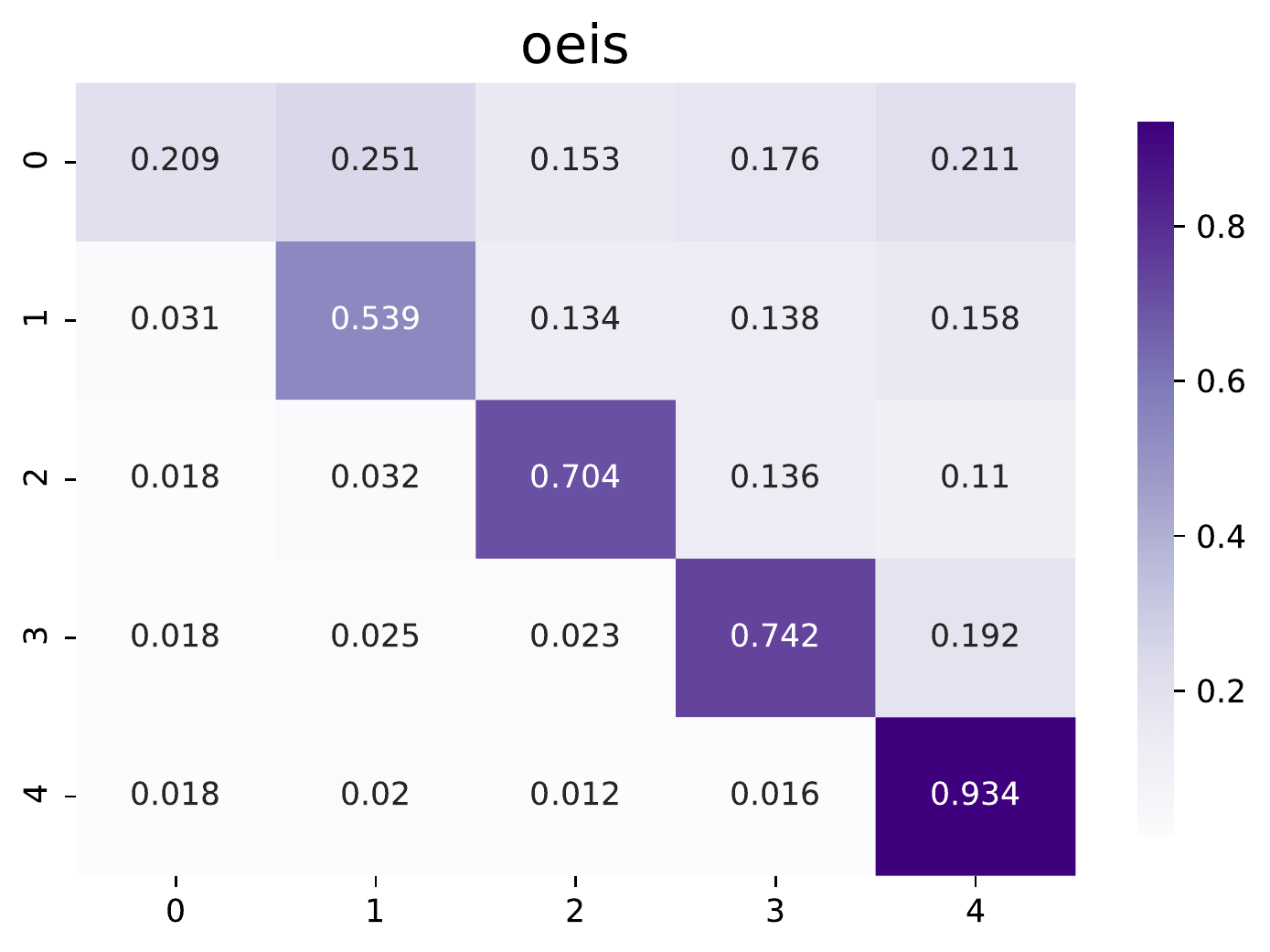}
\end{tabular}
\caption{Compatibility matrices of homophilic datasets. The diagonal dominance indicates strong homophily.}
\label{fig:compat_homophilic}
\end{figure*}

In contrast to the different compatibility matrix structures of our proposed non-homophilous datasets, much other graph data have primarily homophilous relationships, as can be seen in Figure~\ref{fig:compat_homophilic} and Table \ref{tab:homophilic_stats}. The Cora, CiteSeer, PubMed, ogbn-arxiv, and ogbn-products datasets are widely used as benchmarks for node classification \cite{yang2016revisiting, hu2020open}, and are highly homophilous, as can be seen by the diagonally dominant structure of the compatibility matrices and by the high edge homophily and $\hat h$.

We collected the oeis dataset displayed in the bottom right of Figure~\ref{fig:compat_homophilic}. The nodes are entries in the Online Encyclopedia of Integer Sequences \cite{sloane2007line}, and directed edges link an entry to any other entry that it cites. In analogy to arXiv-year and snap-patents, the node labels are the time of posting of the sequence. However, in this case the graph relationships are homophilous, even as we vary the number of distinct classes (time periods). This is in part due to differences between posting in this online encyclopedia and publication of academic papers or patents. For instance, there is less overhead to posting an entry in the OEIS, so users often post separate related entries and variants of these entries in rapid succession. Also, an entry in the encyclopedia often inspires other people to work on similar entries, which can be created in much less time than an academic follow-up work to a given paper. These related entries tend to cite each other, which contributes to homophilic relationships over time. Thus, the data here does not follow the special temporal citation structure of academic publications and patents.

\begin{figure}[ht!]
    \centering
    \includegraphics[width=.35\textwidth]{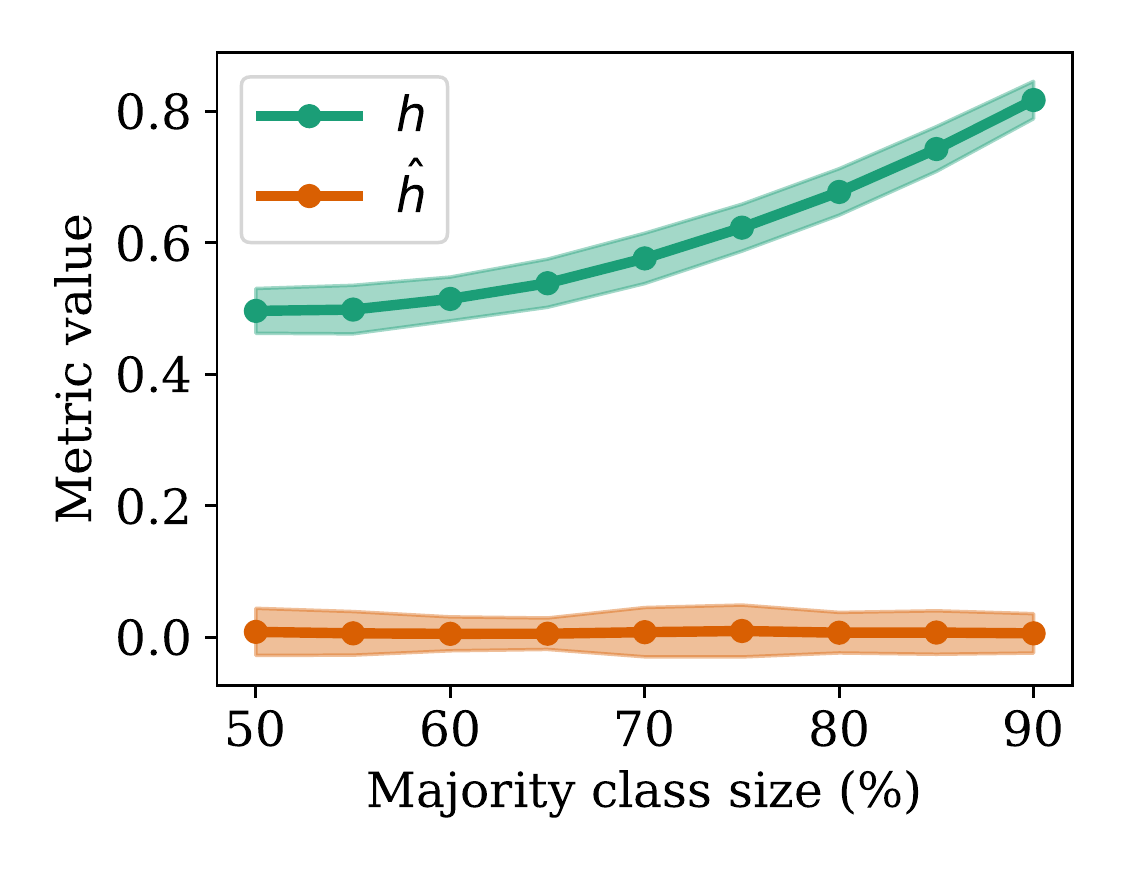}
    \caption{Comparison of edge homophily $h$ and our measure $\hat h$ on random class-imbalanced graph data with edges independent of node labels. Three standard deviations are shaded. Our measure is mostly constant as the classes become more imbalanced, while edge homophily increases.}
    \label{fig:metrics}
\end{figure}

\subsection{Previous Non-Homophilous Data}

\begin{table*}[t]
    \centering
\vspace{10pt}
    {\small
    \begin{tabular}{crrrrrrr}
    \toprule
    Dataset & \# Nodes & \# Edges &  \# Node Feat. & \# C & Context & Edge hom. & $\hat h$ (ours) \\
    \midrule
         Chameleon & 2,277 & 36,101 & 2,325 & 5 & Wiki pages & .23 & .062\\
         Cornell & 183 & 295 & 1,703 & 5 & Web pages & .30 & .047\\
         Actor & 7,600 & 29,926 & 931 & 5 & Actors in movies  & .22 & .011\\
         Squirrel & 5,201 & 216,933 & 2,089 & 5 & Wiki pages & .22 & .025\\
         Texas  & 183 & 309 & 1,703 & 5 & Web pages & .11 & .001\\
         Wisconsin & 251 & 499 & 1,703 & 5 & Web pages & .21 & .094\\
    \bottomrule
    \end{tabular}
    }
    \vspace{5pt}
    \caption{Statistics for datasets from \citet{pei2019geom}. \#C is the number of node classes.}
    \label{tab:geom_gcn}
\end{table*}

\begin{figure*}[t]
\centering
\begin{tabular}{ccc}
    \includegraphics[width=.27\textwidth]{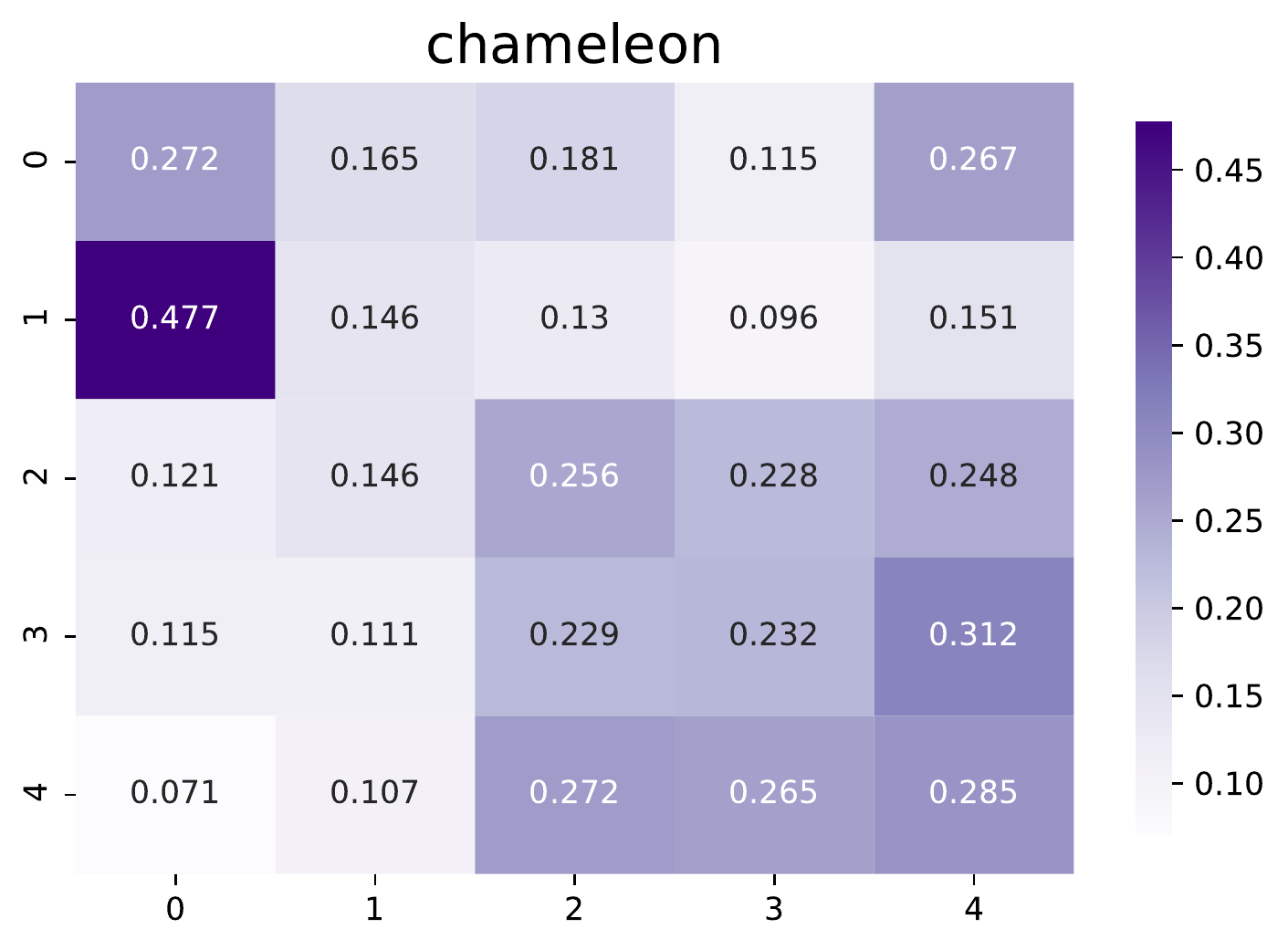} &
    \includegraphics[width=.27\textwidth]{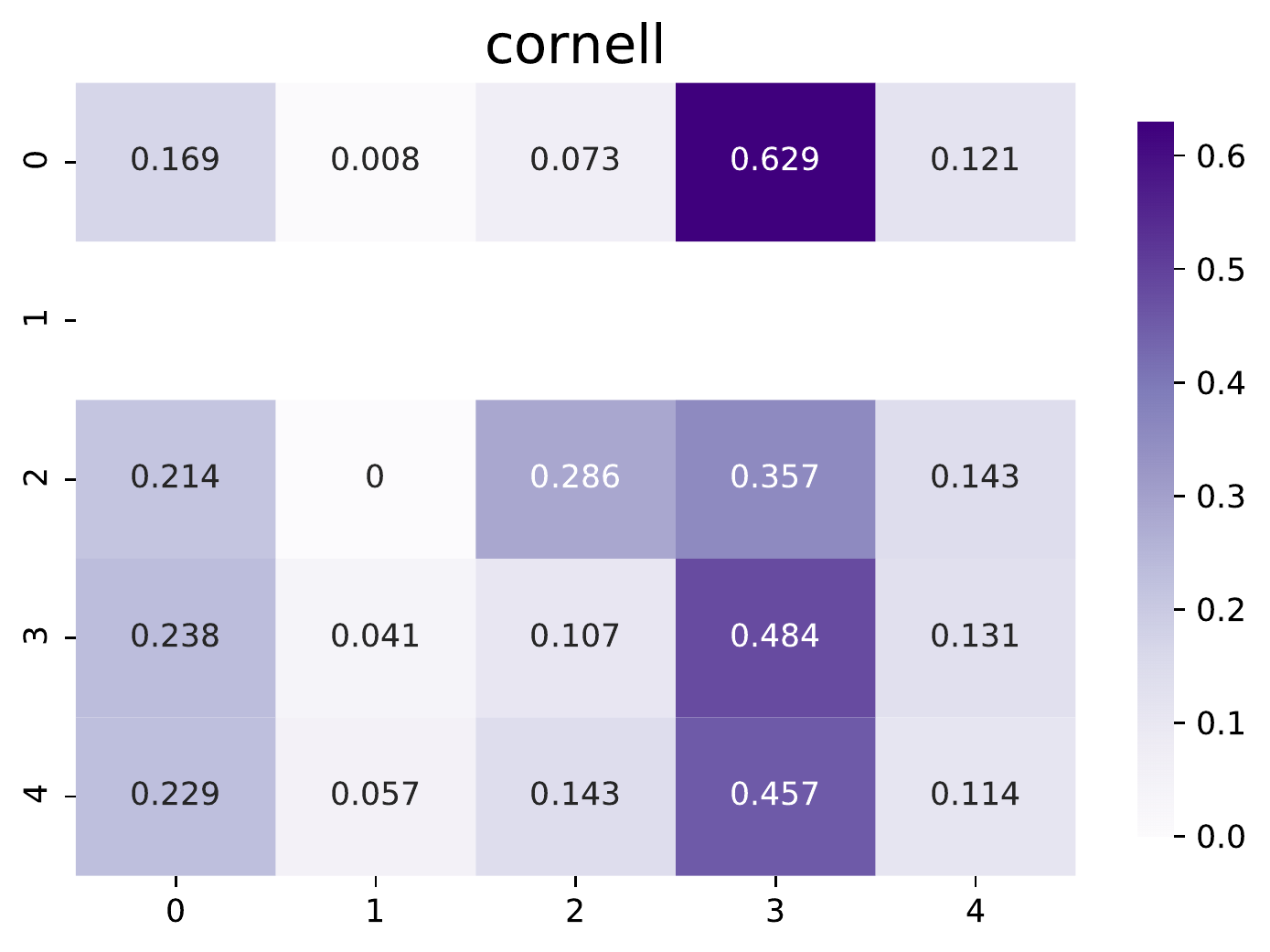} &
    \includegraphics[width=.27\textwidth]{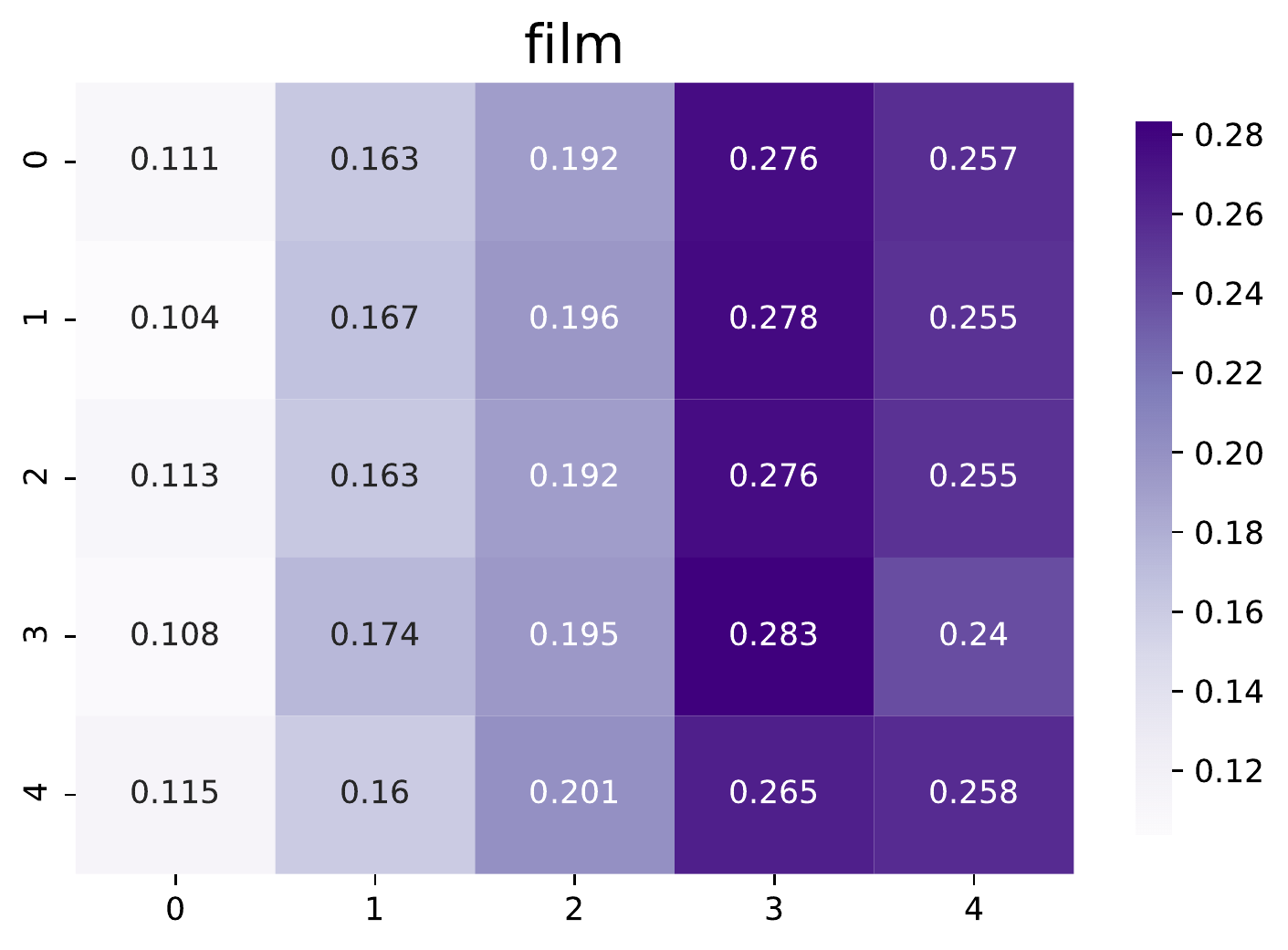} \\
    \includegraphics[width=.27\textwidth]{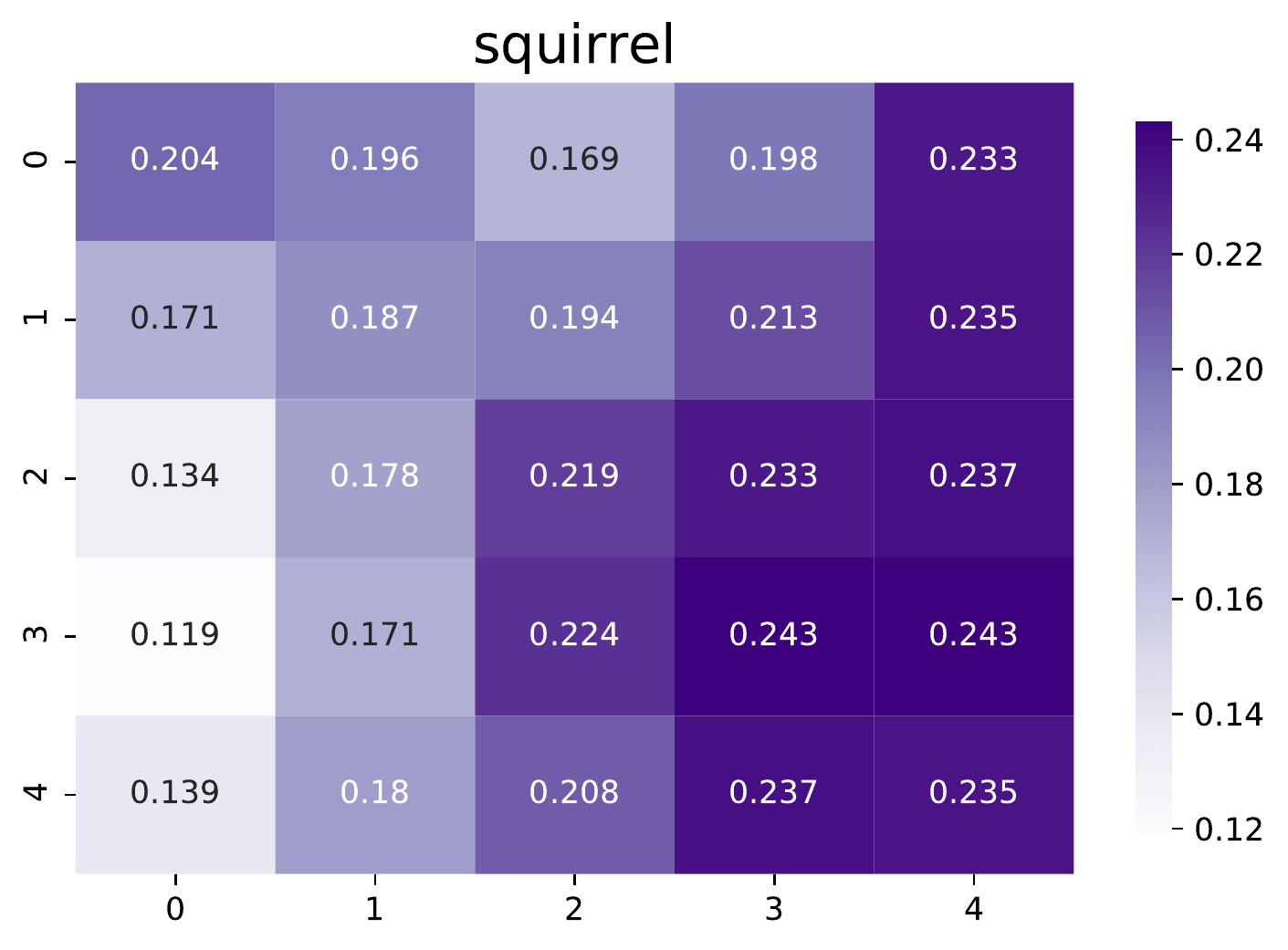} &
    \includegraphics[width=.27\textwidth]{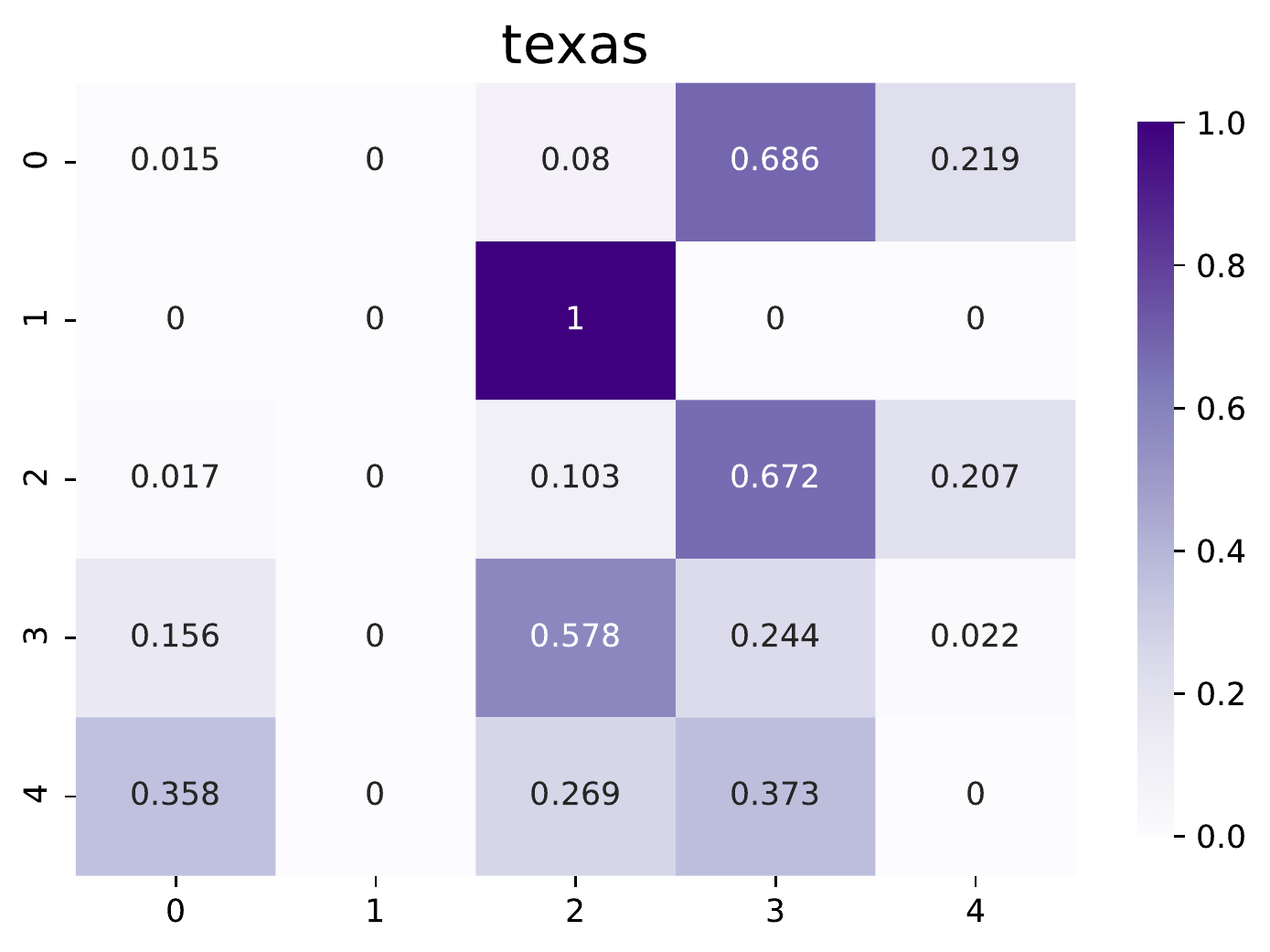} &
    \includegraphics[width=.27\textwidth]{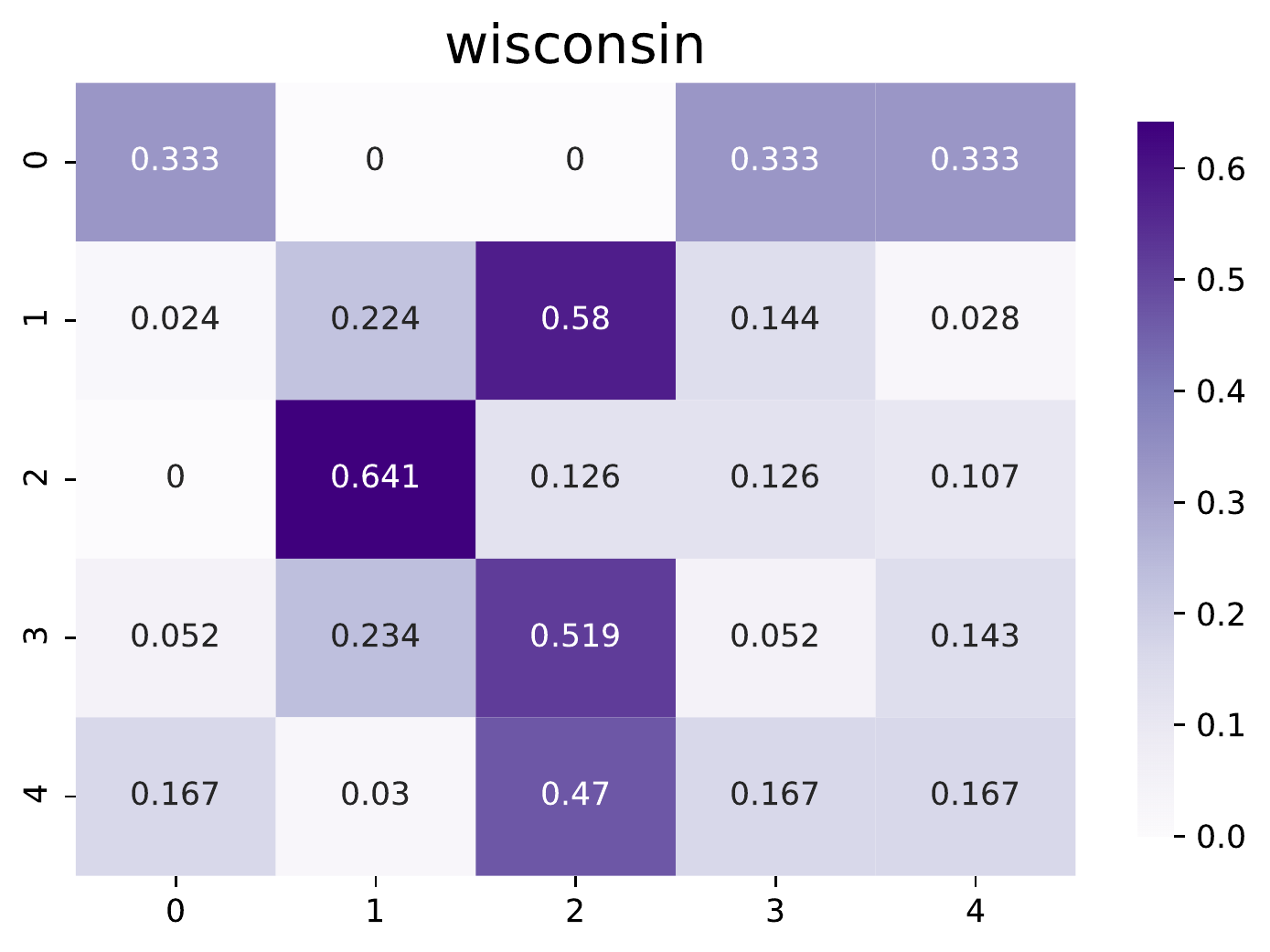}
\end{tabular}
\caption{Compatibility matrices of datasets in \citet{pei2019geom}. The ``film'' dataset is also referred to as ``Actor''. Note that there are no edges leading out of the nodes of class 1 in the Cornell dataset, so there is an empty row in its matrix.}
\label{fig:compat_geom_gcn}
\end{figure*}

For the six datasets in \citet{pei2019geom} often used in evaluation of graph representation learning methods in non-homophilous regimes \cite{zhu2020beyond}, basic statistics are listed in Table \ref{tab:geom_gcn} and compatibility matrices are displayed in Figure \ref{fig:compat_geom_gcn}. We propose many more datasets (including the 100 graphs from Facebook100) that have up to orders of magnitude more nodes and edges and come from a wider range of contexts. There are several cases of class-imbalance in these datasets, which may make the edge homophily misleading. As discussed in Section \ref{sec:measure}, our measure may be able to alleviate issues with edge homophily in measuring homophily of these datasets, and offers a way to distinguish between the Chameleon, Actor, and Squirrel datasets that all have similar edge homophily.

\subsection{Class-Imbalance and Metrics}

In this section, we present experiments that demonstrate an instance in which our metric is not affected by imbalanced classes, while edge homophily is. We generate graphs in which node labels are independent of edges by randomly choosing node labels and generating graph edges by the Erd\H{o}s-R\'enyi random graph model \cite{erdHos1960evolution}. In particular, we fix the number of classes to two, the number of nodes to 100, and the probability of edge formation as .25 between every pair of nodes. Then we generate 100 samples of these random graphs, and compute the mean and standard deviation of both edge homophily $h$ and our measure $\hat h$. As seen in Figure~\ref{fig:metrics}, our measure $\hat h$ is constantly near zero as we increase the size of the majority class, while the edge homophily $h$ increases as the size of majority class increases.

\section{Experimental Details}\label{sec:exp_details}

For gradient-based optimization, we use the AdamW optimizer \cite{kingma2014adam, loshchilov2018decoupled} with weight decay .001  and learning rate $.01$ by default, unless we tune the optimizer for a particular method (as noted below in \ref{sec:hparams}). In all cases, we use full batch gradient descent across the entire graph dataset. Hyperparameter tuning is conducted using grid search for most methods. Tuning for C\&S is done as in the original paper \cite{huang2021combining}, which uses Optuna \cite{optuna} for Bayesian hyperparameter optimization. GCN and GCN-JK on ogbn-proteins use the hyper-parameters of methods on the Open Graph Benchmark leaderboards \cite{hu2020open}. All graphs are treated as undirected besides arXiv-year and snap-patents, in which the directed nature of the edges capture useful temporal information; however, we find that label propagation and C\&S (which builds on label propagation) perform better with undirected graphs in these cases, so we keep the graphs as undirected for these methods. 

Simple methods are run on a Nvidia 2080 Ti with 11 GB GPU memory. In cases where the Nvidia 2080Ti did not provide enough memory, we re-ran experiments on a Nvidia Titan RTX with 24GB GPU memory, reporting (M) in Table \ref{tab:results} if the GPU memory was still insufficient. 
 
\subsection{Hyperparameters}\label{sec:hparams}

Experimental results are reported on the hyperparameter settings below, where we choose the settings that achieve the highest performance on the validation set.
We choose hyperparameter grids that do not necessarily give optimal performance, but hopefully cover enough regimes so that each model is reasonably evaluated on each dataset. Unless otherwise stated, each GNN has dropout of .5 \cite{srivastava2014dropout} and BatchNorm \cite{ioffe2015batch} in each layer. The hyperparameter grids for the different methods are:

\begin{itemize}
    \item MLP: hidden dimension $\in \{16, 32, 64, 128, 256\}$, number of layers $\in \{2,3\}$. We use ReLU activations.
    \item Label propagation: $\alpha \in \{.01, .1, .25, .5, .75, .9, .99\}$. We use 50 propagation iterations.
    \item LINK: weight decay $\in \{.001, .01, .1\}$.
    \item SGC: weight decay $\in \{.001, .01, .1\}$.
    \item C\&S: Normalized adjacency matrix A\textsubscript{1}, A\textsubscript{2} $\in \{D^{-\frac{1}{2}}AD^{-\frac{1}{2}}, \\D^{-1}A, AD^{-1} \}$ for the residual propagation and label propagation, where $A$ is the adjacency matrix of the graph and $D$ is the diagonal degree matrix; $\alpha_1, \alpha_2 \in (0.0, 1.0)$ for the two propagations. Both Autoscale and FDiff-scale were used for all experiments, and scale $\in (0.1, 10.0)$ was searched in FDiff-scale settings. The base predictor is chosen as the best MLP model for each dataset.
    \item GCN: lr $\in \{.1, .01, .001\}$, hidden dimension $\in \{4, 8, 16, 32, 64\}$, except for snap-patents and pokec, where we omit hidden dimension = 64, and ogbn-proteins, where we use the experimental setup of \citet{hu2020open}. Each activation is a ReLU. 2 layers were used for all experiments except for ogbn-proteins.
    \item GAT: lr $\in \{.1, .01, .001\}$. For snap-patents and pokec: hidden channels $\in \{4, 8, 12\}$ and gat heads $\in \{2, 4\}$. For all other datasets: hidden channels $\in \{4, 8, 12, 32\}$ and gat heads $\in \{2, 4, 8\}$. We use the ELU activation \cite{clevert2015fast}. 2 layers were used for all experiments. 
    \item GCN-JK: Identical for GCN, also including JK Type $\in \{\text{cat}, \text{max} \}$.
    \item GAT-JK: Identical for GAT, also including JK Type $\in \{\text{cat}, \text{max} \}$. 
    \item APPNP: MLP hidden dimension $\in \{16, 32, 64, 128, 256\}$, learning rate $\in \{.01, .05, .002\}$, $\alpha \in \{.1, .2, .5, .9\}$. We truncate the series at the $K=10$th power of the adjacency. We also try networks with and without BatchNorm.
    \item H\textsubscript{2}GCN: hidden dimension $\in \{8, 16, 32, 64\}$, number of layers $\in \{1,2\}$, dropout $\in \{0, .5\}$. The architecture follows Section 3.2 of \cite{zhu2020beyond}.
    \item MixHop: hidden dimension $\in \{8, 16, 32\}$, number of layers $\in \{2, 3\}$. Each layer has uses the 0th, 1st, and 2nd powers of the adjacency and has ReLU activations. The last layer is a linear projection layer, instead of the attention output mechanism in \cite{abu2019mixhop}.
    \item GPR-GNN: The basic setup and grid is the same as that of APPNP. We use their Personalized PageRank weight initialization.
\end{itemize}

\section*{Version Notes}

arXiv v2 (July 2021): added batch normalization to the APPNP and GPR-GNN hyperparameter grids, as this sometimes really affects performance --- especially on larger graphs.

\end{document}